\documentclass[pdflatex,sn-mathphys-num]{sn-jnl}

\usepackage{graphicx}%
\usepackage{multirow}%
\usepackage{makecell}%
\usepackage{amsmath,amssymb,amsfonts}%
\usepackage{amsthm}%
\usepackage{mathrsfs}%
\usepackage[title]{appendix}%
\usepackage{xcolor}%
\usepackage{pifont}%
\usepackage{array}%
\usepackage{colortbl}%
\usepackage{nicefrac}%
\usepackage{textcomp}%
\usepackage{manyfoot}%
\usepackage{booktabs}%
\usepackage{algorithm}%
\usepackage{algorithmicx}%
\usepackage{algpseudocode}%
\usepackage{listings}%
\usepackage{bm}%

\theoremstyle{thmstyleone}%
\theoremstyle{thmstyletwo}%

\theoremstyle{thmstylethree}%

\usepackage[capitalize,noabbrev]{cleveref}

\crefname{equation}{Eq.}{Eqs.}
\Crefname{equation}{Equation}{Equations}

\crefname{section}{Sec.}{Secs.}
\Crefname{section}{Section}{Sections}

\crefname{table}{Table}{Tables}
\Crefname{table}{Table}{Tables}

\crefname{algorithm}{Algorithm}{Algorithms}
\Crefname{algorithm}{Algorithm}{Algorithms}

\newcommand{\safeincludegraphics}[2][]{%
  \IfFileExists{#2}{%
    \includegraphics[#1]{#2}%
  }{%
    \fbox{\parbox[c][0.18\textheight][c]{0.88\linewidth}{\centering Missing figure:\\\ttfamily\detokenize{#2}}}%
  }%
}

\newcommand{\be}{\begin{eqnarray} \begin{aligned}}
\newcommand{\ee}{\end{aligned} \end{eqnarray} }
\newcommand{\benn}{\begin{eqnarray*} \begin{aligned}}
\newcommand{\eenn}{\end{aligned} \end{eqnarray*} }

\usepackage{color}

\crefformat{figure}{fig.~#2#1#3}
\Crefformat{figure}{Fig.~#2#1#3}

\begin{document}

\title[Frequency-Enhanced Diffusion Models: Curriculum-Guided Semantic Alignment for Zero-Shot Skeleton Action Recognition]{Frequency-Enhanced Diffusion Models: Curriculum-Guided Semantic Alignment for Zero-Shot Skeleton Action Recognition}


\author[1]{\fnm{Yuxi} \sur{Zhou}}
\equalcont{These authors contributed equally to this work.}

\author[2]{\fnm{Zhengbo} \sur{Zhang}}
\equalcont{These authors contributed equally to this work, are co-first authors.}

\author[3]{\fnm{Jingyu} \sur{Pan}}

\author[4]{\fnm{Zhiyu} \sur{Lin}}

\author*[1,5]{\fnm{Zhigang} \sur{Tu}}\email{tuzhigang@whu.edu.cn}

\affil[1]{\orgdiv{State Key Laboratory of Information Engineering in
Surveying, Mapping and Remote Sensing}, \orgname{Wuhan University}, \orgaddress{ \city{Wuhan}, \country{China}}}
\affil[2]{\orgdiv{Information Systems Technology and Design Pillar}, \orgname{Singapore University of Technology and Design}, \orgaddress{ \city{Singapore}, \country{Singapore}}}
\affil[3]{\orgdiv{School of Geodesy and Geomatics}, \orgname{Wuhan University}, \orgaddress{ \city{Wuhan}, \country{China}}}
\affil[3]{\orgdiv{School of Mathematics and Statistics}, \orgname{Wuhan University}, \orgaddress{ \city{Wuhan}, \country{China}}}

\affil[5]{\orgdiv{Wuhan University Shenzhen Research Institute}, \orgaddress{ \city{Shenzhen}, \country{China}}}




\abstract{
Human action recognition is pivotal in computer vision, with applications ranging from surveillance to human-robot interaction. Despite the effectiveness of supervised skeleton-based methods, their reliance on exhaustive annotation limits generalization to novel actions. Zero-Shot Skeleton Action Recognition (ZSAR) emerges as a promising paradigm, yet it faces challenges due to the spectral bias of diffusion models, which oversmooth high-frequency dynamics. Here, we propose Frequency-Aware Diffusion for Skeleton-Text Matching (FDSM), integrating a Semantic-Guided Spectral Residual Module, a Timestep-Adaptive Spectral Loss, and Curriculum-based Semantic Abstraction to address these challenges. Our approach effectively recovers fine-grained motion details, achieving state-of-the-art performance on NTU RGB+D, PKU-MMD, and Kinetics-skeleton datasets. Code has been made available at \url{https://github.com/yuzhi535/FDSM}. Project homepage: \url{https://yuzhi535.github.io/FDSM.github.io/}
}




\pagenumbering{arabic}
\maketitle

\section{Introduction}\label{sec:intro}
Human action recognition stands as a cornerstone of computer vision, underpinning applications ranging from intelligent surveillance~\cite{singh2019multi, xia2026uavovooutofviewpointgeneralizationuav} to human-robot interaction~\cite{singh2019multi, hong2025advanced}. While early approaches predominantly relied on RGB video data, the field has witnessed a paradigm shift toward skeleton-based modalities, driven by the proliferation of cost-effective depth sensors and robust pose estimation algorithms~\cite{yang2025end, aouaidjia2019efficient, hu2023human}. 
However, despite the efficacy of fully supervised skeleton-based methods~\cite {hou2016skeleton, chi2022infogcn, qiu2023effective, liu2023transkeleton, wu2024frequency, zhao2025dual, xie2025enhanced}, 
their reliance on exhaustive annotation limits generalization to fixed categories, rendering them ineffective for open-world scenarios where novel actions emerge continuously~\cite{tu2025informative}.
To overcome this bottleneck, Zero-Shot Skeleton Action Recognition (ZSAR) has emerged as a compelling paradigm. By exploiting auxiliary semantic information—ranging from manual attributes for seen classes to large-scale pre-trained language knowledge for unseen concepts—ZSAR effectively transfers recognition capabilities across the domain gap~\cite{do2025bridging, hubert2017learning, gupta2021syntactically, zhou2023zero,li2023multi,chen2024fine, zhu2024part,li2025sa}. 

Nevertheless, effective ZSAR is hindered by the profound \textbf{modality gap} between the high-frequency spatio-temporal dynamics of skeletal data and the abstract, static nature of semantic descriptions.
Early approaches utilized static discriminative mappings (e.g., VAEs or CLIP) to bridge this divide, yet such rigid point-to-point alignments often struggle to capture the complex temporal evolution of motion~\cite{zhu2024part,do2025bridging}.
Consequently, recent research has pivoted toward generative paradigms, specifically diffusion models~\cite{ho2020denoising, Rombach22LDM, do2025bridging,zhang2024diff}.
Unlike discriminative methods,
diffusion models utilizes the reverse diffusion process to implicitly align skeleton and
text features within a shared latent space, ensuring that generated outputs adhere closely to the given condition~\cite{do2025bridging}. By modeling the data distribution as a gradual denoising process, diffusion models demonstrate a remarkable capacity to capture the intricate manifold of skeletal motion, overcoming the challenges of direct feature space alignment between skeleton and text modalities.

\begin{figure}[t]
    \centering
    \safeincludegraphics[width=0.9\linewidth]{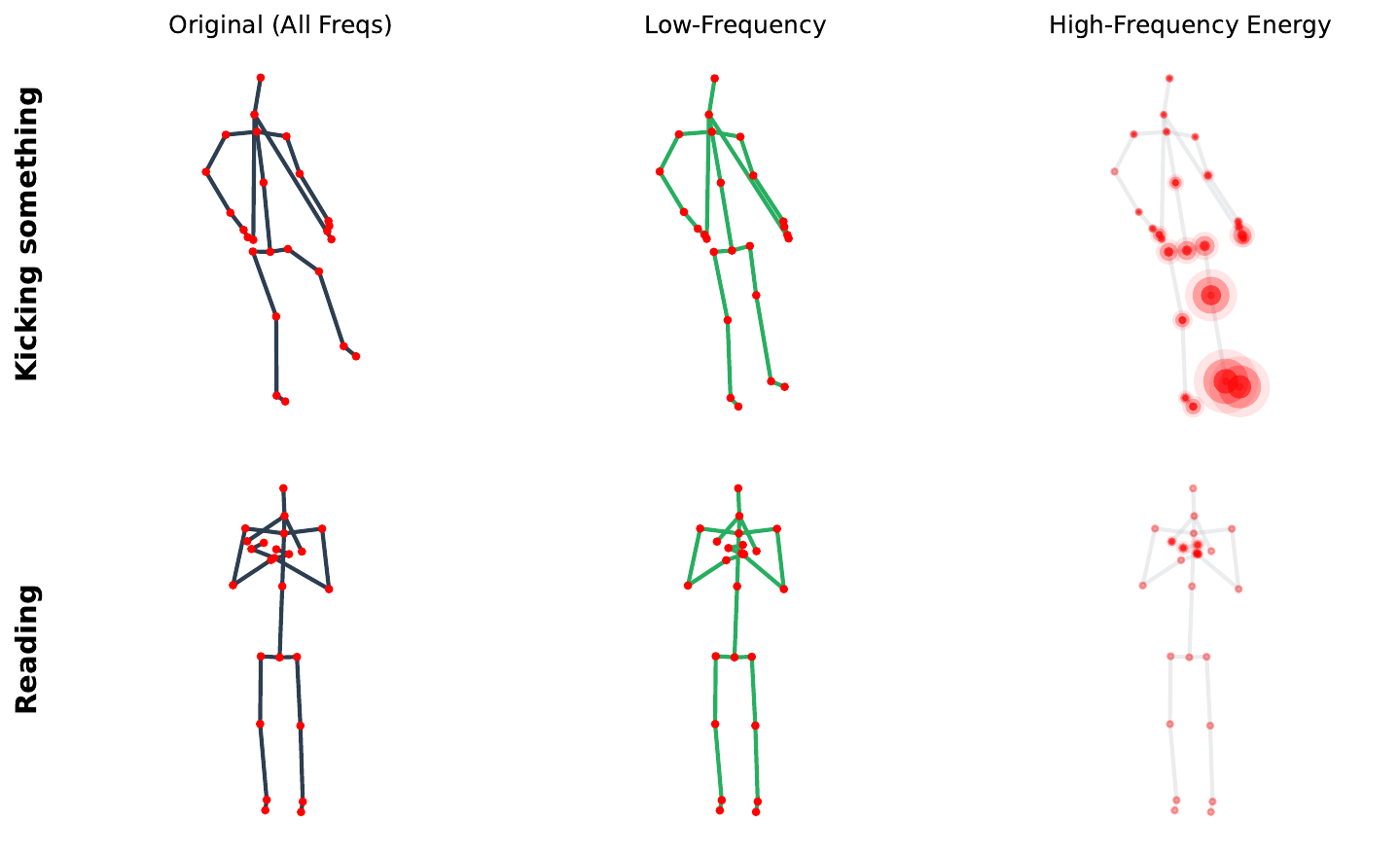}
    \caption{Frequency-domain visualization of skeleton sequences via DCT decomposition. Each row shows one action: (left) original skeleton with all frequency components; (center) low-frequency reconstruction capturing global pose topology; (right) high-frequency energy heat map, where bubble size at each joint indicates the magnitude of residual components. High-frequency micro-dynamics concentrate at extremities (hands, feet) in dynamic actions (e.g., Clapping, Waving hand) but are absent in near-static actions (e.g., Reading), illustrating why high-frequency suppression in the Diffusion Transformer degrades fine-grained action discrimination.}
    \label{fig:heatmap}
\end{figure}

However, despite this theoretical promise, the direct application of standard diffusion paradigms to ZSAR exposes critical vulnerabilities rooted in the interplay between signal frequency and semantic interpretation. 
\textbf{Firstly}, regarding frequency, skeletal actions are spectrally diverse, comprising low-frequency structural components (representing global pose trajectories) and high-frequency details (encoding rapid, fine-grained dynamics)~\cite{wu2025frequency, xia2026efsidetrefficientfrequencysemanticintegration}. As illustrated in Fig.~\ref{fig:heatmap}, we decompose ground-truth skeleton sequences using the DCT along the temporal axis and visualize three components for each action: (1)~the original skeleton with all frequency components; (2)~the low-frequency reconstruction retaining only coefficients $k < L/4$, which captures global pose topology; and (3)~a high-frequency energy heat map, where bubble size at each joint indicates the magnitude of the high-frequency residual ($k \geq L/4$). The contrast is clear: for dynamic actions such as \textit{Kicking something} or \textit{Clapping}, large bubbles concentrate at the foot, leg, and hand joints, reflecting the rapid micro-dynamics that are critical for distinguishing these classes. For near-static actions such as \textit{Reading}, bubbles are uniformly small, confirming that the low-frequency reconstruction already captures the relevant motion. This visualization directly demonstrates that high-frequency components are action-discriminative and must not be suppressed—yet the standard Diffusion Transformer does precisely that. However, standard diffusion models face a \textit{dual spectral bottleneck} that independently constrains high-frequency synthesis. On the architectural side, recent theoretical findings~\cite{si2022inception} reveal that Transformers, empirically identified as the optimal backbone for diffusion-based ZSAR~\cite{do2025bridging}, exhibit an inherent ``low-pass'' inductive bias, causing the backbone to act as a spectral filter that smooths out sharp, rapid motion cues essential for distinguishing complex actions. On the optimization side, standard Mean Squared Error objectives inherently average out stochastic high-frequency variations, while training dynamics favor low-frequency convergence~\cite{zhang2018unreasonable, rahaman2019spectral}. Together, these dual biases result in generated motions that are structurally coherent but dynamically oversmoothed, lacking the discriminative high-frequency fidelity required for zero-shot recognition.\textbf{Secondly}, regarding semantics, current methods~\cite{do2025bridging} face a semantic-dynamic ambiguity. Conditioning generation solely on coarse-grained class labels (e.g., ``Walking'') fails to explicitly convey the fine-grained physical dynamics—such as tempo, rhythm, and intensity—that define the motion's spectral signature. Without this explicit dynamic guidance, the model struggles to determine whether high-frequency components in the target data represent essential action details (as in ``Punching/Slapping'') or mere noise (as in ``Sitting down''), leading to synthesized features that are either structurally oversmoothed or polluted with hallucinations.

To address these limitations, we construct a unified \textbf{Frequency-Aware Diffusion for Skeleton-Text Matching (FDSM)} framework that establishes coherent spectral constraints between the input condition and the optimization objective. Our approach is built upon three synergistic components. To counteract the backbone's spectral bias, we introduce a frequency-domain residual module that selectively amplifies high-frequency dynamics based on kinematic priors. Complementing this, we design a dynamic spectral loss that aligns frequency supervision with the diffusion model's native coarse-to-fine generation process, ensuring physically valid reconstruction without overfitting to noise. Furthermore, to bridge the semantic gap, we employ a curriculum learning strategy that transfers rich kinematic knowledge from LLMs into the visual encoder, enabling robust inference from sparse labels.

Our main contributions can be summarized as follows:
\begin{itemize}
\item We propose the \textbf{Semantic-Guided Spectral Residual Module} to address the inherent ``low-pass'' bias of standard generative backbones. By establishing a frequency-domain residual pathway, this module explicitly amplifies discriminative high-frequency dynamics, effectively counteracting over-smoothing and enabling the synthesis of fine-grained motion details.
\item We introduce the \textbf{Timestep-Adaptive Spectral Loss} as a synergistic optimization objective. This mechanism aligns the frequency supervision with the diffusion model's denoising schedule, preventing overfitting to noise in early stages while enforcing rigorous high-frequency reconstruction in the final refinement steps.
\item We design a \textbf{Curriculum-based Semantic Abstraction} strategy to bridge the semantic gap in zero-shot inference. By training with a `rich-to-sparse' curriculum of kinematic descriptions, we force the visual encoder to internalize complex motion priors, enabling robust generalization from sparse labels.
\item Extensive experiments on three benchmark datasets (NTU RGB+D~\cite{Ntu60}, NTU RGB+D 120~\cite{Ntu120},  PKU-MMD~\cite{PKUMMD} and Kinetics-skeleton~\cite{yan2018spatial,Kinetics}) demonstrate that our method achieves state-of-the-art performance, significantly outperforming existing generative and discriminative baselines in both Zero-Shot (ZSL) and Generalized Zero-Shot (GZSL) settings.
\end{itemize}



\section{Related Work}\label{sec:related_work}

\subsection{Zero-Shot Skeleton Action Recognition}
The fundamental goal of Zero-Shot Skeleton Action Recognition (ZSAR) is to identify human behaviors from novel categories without access to labeled training instances. The dominant paradigm addresses the heterogeneity between skeletal kinematics and textual semantics by constructing a shared embedding space. Contemporary literature classifies these efforts into three primary streams: Variational Autoencoder (VAE)-based frameworks~\cite{CADAVAE, SynSE, MSF, SADVAE, wu2025frequency}, Contrastive Learning-based approaches~\cite{SMIE, PURLS, STAR, DVTA, InfoCPL}, and emerging Diffusion-based methods~\cite{do2025bridging}.

\textbf{VAE-based Approaches.} Pioneering works such as CADA-VAE~\cite{CADAVAE} employ cross-modal VAEs to align latent distributions, enforcing cycle-consistency constraints where each modality's decoder reconstructs features from the other's latent code. SynSE~\cite{SynSE} refines this by adopting a decoupled generative strategy, training separate VAEs for verbs and nouns to form a structured semantic manifold. To enhance semantic granularity, MSF~\cite{MSF} incorporates multi-level descriptions, synthesizing action-level labels with motion-level details. Furthermore, addressing inherent skeletal noise, SA-DVAE~\cite{SADVAE} introduces a disentanglement mechanism to isolate semantically relevant features from extraneous variations, ensuring alignment between text embeddings and informative skeletal components. More recently, FS-VAE~\cite{wu2025frequency} extends the VAE paradigm with DCT-based frequency-semantic enhancement, hierarchical motion descriptions, and a calibrated cross-alignment loss, showing that frequency-domain cues carry discriminative information beyond global pose topology. However, this line of work focuses on representation alignment and does not address the spectral bias introduced by the denoising process in diffusion-based frameworks.

\textbf{Contrastive Learning-based Approaches.} These methods prioritize cross-modal consistency through contrastive objectives~\cite{SimCLR}. SMIE~\cite{SMIE} integrates skeletal and textual features using a masking strategy, treating occluded parts as positive samples to contrast against inter-class negatives. PURLS~\cite{PURLS} harnesses Large Language Models (LLMs) like GPT-3~\cite{GPT} to generate detailed descriptions of body part evolution, employing cross-attention to guide visual-semantic alignment. Building on this, STAR~\cite{STAR} utilizes GPT-3.5~\cite{GPT} to produce hierarchical descriptions for anatomical groups, introducing learnable prompts to refine matching. DVTA~\cite{DVTA} proposes a dual-alignment strategy, simultaneously optimizing global feature matching and local cross-attention. Similarly, InfoCPL~\cite{InfoCPL} enriches the semantic space by generating extensive sentence variations per action, thereby strengthening the contrastive learning manifold.

\textbf{Diffusion-based Approaches.} Departing from rigid discriminative alignments, recent research has shifted towards generative diffusion models. While primarily explored in text-to-motion generation tasks \cite{MDM, MotionDiffuse}, the diffusion paradigm has recently been adapted to the zero-shot recognition setting. Pioneering this direction, TDSM \cite{do2025bridging} introduced a conditional diffusion framework to synthesize skeleton features from noise, guiding the denoising trajectory of skeletal features using semantic contexts, theoretically achieving implicit modal harmonization and improved generalization. However, existing methods often overlook the inherent spectral bias of generative backbones, leading to the over-smoothing of high-frequency components, such as subtle hand movements, which are critical for fine-grained discrimination. In contrast, our framework addresses this limitation by introducing frequency-aware mechanisms that preserve high-frequency fidelity, ensuring the synthesis of skeleton features that are both semantically aligned and dynamically detailed.

\subsection{Skeleton-based Action Recognition}

Unlike zero-shot settings that aim to recognize unseen classes without labeled instances, traditional skeleton-based action recognition operates under fully supervised protocols. Initial approaches \cite{VALSTM, STLSTM} utilized Recurrent Neural Networks (RNNs) to model the temporal evolution of skeletal sequences. Subsequently, Convolutional Neural Networks (CNNs) \cite{Ske2Grid, PoseC3D} were explored, transforming skeletal data into pseudo-image representations. More recently, Graph Convolutional Networks (GCNs) \cite{CTRGCN, InfoGCN, MSG3D, BlockGCN, XIA2024107210} have become the dominant paradigm, effectively capturing the topological structure of joints and bones. Pioneering this direction, ST-GCN \cite{STGCN} introduced spatial-temporal graph convolutions, while Shift-GCN \cite{ShiftGCN} significantly enhanced computational efficiency through shift graph operations. To overcome the receptive field limitations of GCNs, Transformer-based architectures \cite{IGFormer, SkateFormer, wu2024frequency, zhao2026multi, sun2024decoupled, zhang2023graph, yao2023scene} have been proposed to model global dependencies. In this study, we employ the established ST-GCN \cite{STGCN} and Shift-GCN \cite{ShiftGCN} backbones to extract robust skeletal-temporal representations, mapping input sequences into a latent space for subsequent processing.

\subsection{Diffusion Models}

Diffusion models have revolutionized generative tasks by learning to invert progressive noise corruption, enabling the reconstruction of complex data distributions. Denoising Diffusion Probabilistic Models (DDPMs)~\cite{DDPM} established this sequential paradigm, modeling intricate manifolds through iterative refinement. To alleviate computational costs, Latent Diffusion Models (LDMs)~\cite{SD} perform generation within a compressed latent space, balancing efficiency with high-fidelity output. The efficacy of LDMs in cross-modal tasks (e.g., text-to-image synthesis~\cite{SD} and text-to-motion inbetweening~\cite{peng2025precise}) highlights their potential for multimodal synchronization, typically employing U-Net backbones~\cite{Unet} with cross-attention to inject semantic guidance. Recently, Diffusion Transformers (DiTs)~\cite{DiT} have advanced this architecture by integrating scalable transformer blocks directly into the diffusion process. In this work, following the paradigm established by~\cite{do2025bridging}, we leverage the generative synergy of diffusion models not merely for synthesis but for robust feature alignment. Specifically, we utilize a DiT-based backbone as a denoising engine, where textual descriptions condition the restoration of noisy skeletal features. This mechanism effectively anchors semantic knowledge within the latent space, fostering a resilient cross-modal alignment essential for zero-shot generalization.

\begin{figure}
    \centering
    \includegraphics[width=1.12\linewidth, trim={0.9cm 0 0 0}, clip]{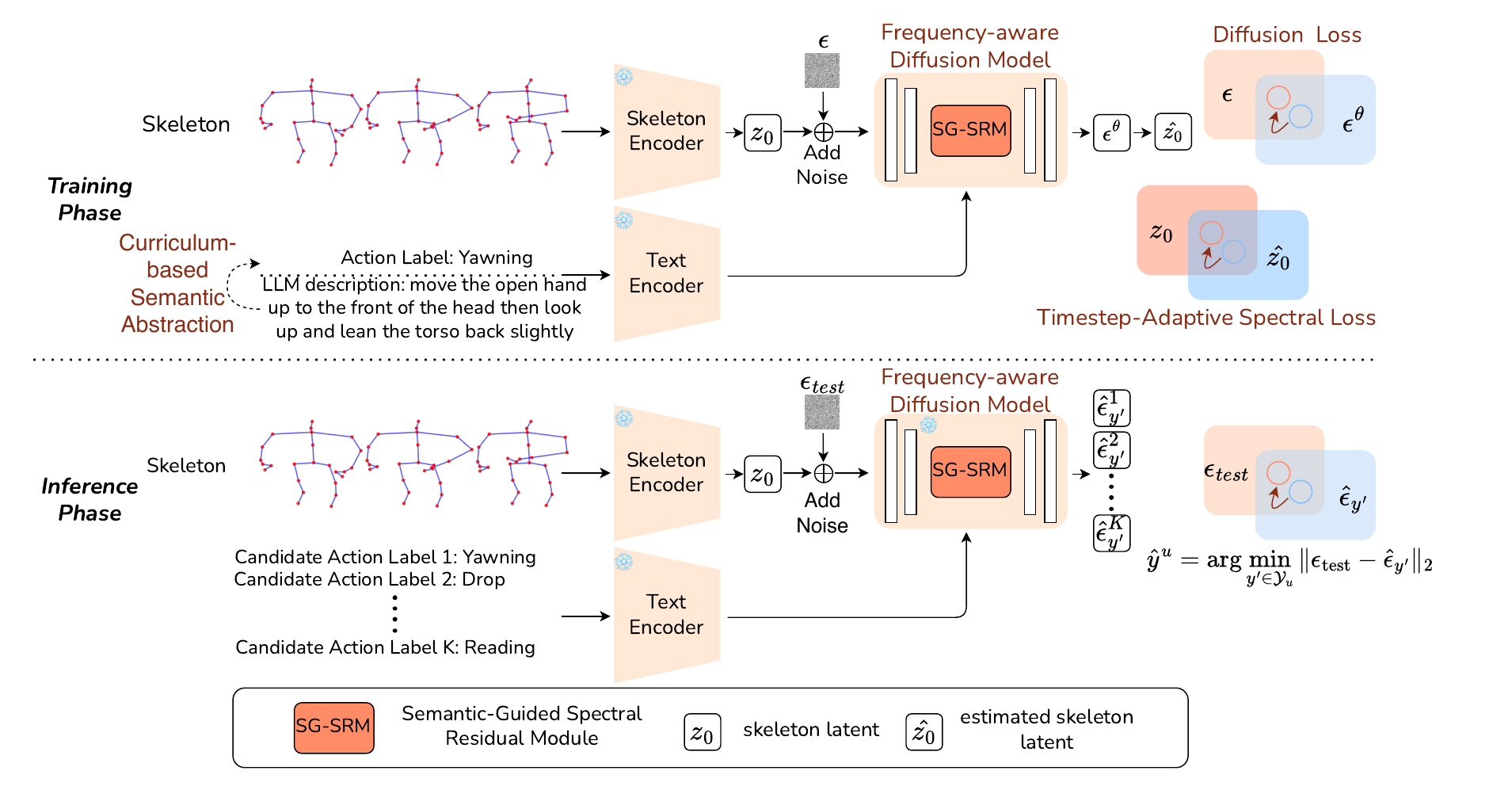}
    \caption{
    Overview of the proposed Frequency-Aware Diffusion Framework. The system integrates three synergistic modules: (1) Semantic-Guided Spectral Residual Module (SG-SRM) to amplify high-frequency dynamics based on LLM-derived kinematic priors; (2) Timestep-Adaptive Spectral Loss to enforce coherent spectral supervision during denoising; and (3) Curriculum-based Semantic Abstraction to bridge the semantic gap via a `rich-to-sparse' training strategy.
    }
    \label{fig:pipeline}
\end{figure}

\section{Method}\label{sec:method}

We present \textbf{Frequency-Aware Diffusion for Skeleton-Text Matching (FDSM)}, a unified framework designed to reconcile the inherent spectral limitations of generative backbones with the stochastic nature of skeletal data. Our approach stems from the critical insight that standard Diffusion Transformers exhibit a strong ``spectral bias,'' effectively functioning as low-pass filters that preserve global pose structure but aggressively attenuate the high-frequency dynamics essential for motion realism. To overcome this inductive bias without overfitting to sensor noise, we construct a holistic frequency regulation mechanism governed by the interplay of semantic capacity and spectral optimization constraints.

As showed in~\cref{fig:pipeline}, the framework is actualized through three synergistic technical contributions. 
\textbf{First}, we introduce the {Semantic-Guided Spectral Residual Module} to compensate for the backbone's architectural spectral bias. Functioning as a dynamic gain controller in the frequency domain, this module selectively amplifies high-frequency magnitudes via a DCT-IDCT pathway. Crucially, the amplification is gated by a predicted kinematic intensity score, which is generated by an internal projection head {distilled} from LLM knowledge, ensuring signal enhancement is semantically warranted.
\textbf{Second}, we impose the {Timestep-Adaptive Spectral Loss} to strictly supervise the synthesis process. Recognizing the intrinsic {coarse-to-fine} generative trajectory of diffusion models, this objective dynamically modulates frequency supervision based on the denoising timestep—suppressing high-frequency penalties during early noise-dominated stages and progressively releasing them for final refinement.
\textbf{Finally}, to resolve the semantic ambiguity where sparse labels fail to convey complex motion patterns, we propose {Curriculum-based Semantic Abstraction}. By training with a `rich-to-sparse' curriculum of descriptive prompts, we force the visual encoder to internalize {structural kinematic correlations} (e.g., temporal phases and coordination), enabling robust zero-shot inference where the model can infer correct frequency distributions even from minimal textual cues.

\subsection{Semantic-Guided Spectral Residual Module}
\label{sec:module}

As discussed above, standard diffusion transformers exhibit an inherent ``spectral bias'', acting as low-pass filters that preserve global pose structures while attenuating fine-grained motion dynamics~\cite{Chen2022VisionTA, si2022inception}. 
To compensate for this limitation, we introduce the \textbf{Semantic-Guided Spectral Residual Module (SG-SRM)}. Unlike standard attention mechanisms that operate in the spatial-temporal domain, SG-SRM functions as a distinct frequency-domain gain controller. It explicitly amplifies high-frequency components to restore motion sharpness, but crucially, this amplification is gated by a semantic prior to prevent the over-enhancement of sensor noise.

\vspace{1mm}
\noindent \textbf{Frequency-Domain Transformation.}
Following~\cite{do2025bridging}, we operate on the latent space extracted by the pre-trained encoder. 
To address the backbone's spectral bias, we must explicitly disentangle the fine-grained motion dynamics (e.g., speed, rhythm, and jitter) from the global pose structure. We observe that these dynamic attributes are inherently encoded in the temporal evolution of the skeleton, whereas the spatial dimension ($V$) primarily encodes topological graph constraints~\cite{STGCN}. Processing the spatial dimension would disrupt limb connectivity; conversely, decomposing the temporal dimension ($L$) allows us to isolate kinematic patterns based on their rate of change.
Therefore, we perform the Discrete Cosine Transform (DCT) specifically along the temporal dimension $L$. 
This operation converts implicit time-series variations into an explicit spectral distribution, where high-frequency coefficients directly correspond to rapid micro-movements (or sensor noise)~\cite{wu2024frequency,wu2025frequency}. Formally, operating on the latent features $\mathbf{Z} \in \mathbb{R}^{B \times C \times L \times V}$, the spectral coefficient $\mathbf{F}_{b,c,k,v}$ is computed as~\cite{ahmed1974discrete}:
\begin{equation}\label{eq:dct}
    \mathbf{F}_{b,c,k,v} = \beta_k \sum_{l=0}^{L-1} \mathbf{Z}_{b,c,l,v} \cos\left( \frac{\pi (2l+1)k}{2L} \right)
\end{equation}
where $k \in \{0, \dots, L-1\}$ denotes the frequency index, and $\beta_k$ represents the normalization coefficient required to ensure orthogonality, defined as $\beta_0 = \sqrt{1/L}$ and $\beta_k = \sqrt{2/L}$ for $k > 0$.
Crucially, this spectral decomposition unmasks the backbone's limitations~\cite{rahaman2019spectral}: it reveals precisely that the high-frequency coefficients lack sufficient magnitude compared to the natural motion distribution.

\begin{figure}
    \centering
    \includegraphics[width=0.9\linewidth]{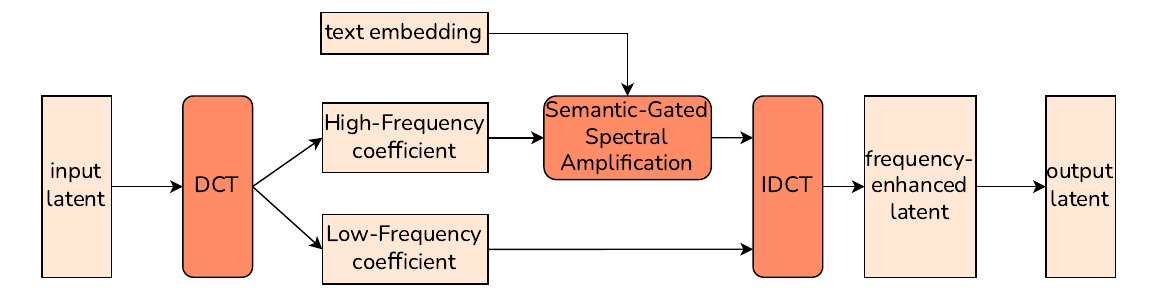}
    \caption{
    Illustration of the Semantic-Guided Spectral Residual Module (SG-SRM). The module performs a DCT along the temporal dimension to decompose latent features into frequency coefficients. A semantic-guided gain filter, derived from the predicted kinematic intensity score, selectively amplifies high-frequency bands. The modulated spectrum is then reverted back to the temporal domain via IDCT, yielding enhanced latent features with restored fine-grained dynamics.
    }
    \label{fig:sgsrm}
\end{figure}

\vspace{1mm}
\noindent \textbf{Semantic-Gated Spectral Amplification.}
Targeting these attenuated frequency bands, our objective is to rectify the spectral distribution distorted by the backbone's low-pass bias. Since the diffusion process disproportionately attenuates the magnitude of high-frequency coefficients, we model the enhancement as a \textit{spectral magnitude restoration process}. \textbf{Specifically}, we construct the modulated spectrum $\tilde{\mathbf{F}}$ via a linear re-weighting mechanism, where $\mathbf{G}_k$ serves as a frequency-specific gain filter:
\begin{equation}\label{eq:spectral_gain}
    \tilde{\mathbf{F}}_{b,c,k,v} = \mathbf{F}_{b,c,k,v} \cdot (1 + \alpha \cdot \mathbf{G}_k)
\end{equation}
Mathematically, the term $(1 + \alpha \cdot \mathbf{G}_k)$ functions as a scalar gain that counteracts the dampening effect of the backbone. By scaling up the magnitude $\mathbf{F}$, we explicitly recover the high-frequency components lost during latent processing, which corresponds to sharpening temporal gradients in the time domain.
This mechanism implements a \textbf{High-Frequency Boosting} strategy, conceptually similar to Unsharp Masking (USM) in signal processing~\cite{polesel2000image}. By formulating the gain as a residual term, the identity term $1$ preserves fundamental structural information, while the dynamic term $\alpha \cdot \mathbf{G}_k$ selectively amplifies suppressed high-frequency components. 
\textbf{Subsequently}, to revert the features back to the temporal domain for subsequent network processing, we perform the Inverse Discrete Cosine Transform (IDCT):
\begin{equation}\label{eq:idct}
    \tilde{\mathbf{Z}}_{b,c,l,v} = \sum_{k=0}^{L-1} \beta_k \tilde{\mathbf{F}}_{b,c,k,v} \cos\left( \frac{\pi (2l+1)k}{2L} \right)
\end{equation}
This operation yields the enhanced latent feature $\tilde{\mathbf{Z}}$. By incorporating the identity term within the spectral gain (\cref{eq:spectral_gain}), the module inherently functions as a frequency-domain residual pathway, where the fundamental structural signal is preserved while discriminative high-frequency components are selectively boosted. Due to the linearity of the transform, this spectral modulation directly translates to sharpening the temporal gradients (motion speed) in the time domain, effectively restoring the fine-grained dynamics suppressed by the backbone.

However, a uniform scaling is theoretically ill-posed due to the uneven Signal-to-Noise Ratio (SNR). High-frequency bands contain both sharp motion details (signal) and sensor jitter (noise). To resolve this ambiguity, we design the gain filter $\mathbf{G}$ based on the \textbf{predicted semantic intensity} $\hat{s}_y$:
\begin{equation}\label{eq:gain_filter}
    \mathbf{G}_k = 
    \begin{cases} 
        0 & \text{if } k < M \quad (\text{Topology Preservation}) \\
        \hat{s}_y & \text{if } k \ge M \quad (\text{Semantic-Aware Gain})
    \end{cases}
\end{equation}
This split-spectrum design addresses two constraints. First, for low frequencies ($k < M$), which encode fundamental pose topology, we enforce $\mathbf{G}_k=0$ to strictly preserve the underlying manifold structure. Second, for high frequencies ($k \ge M$), we use $\hat{s}_y$ as a \textit{contextual switch}. For dynamic actions ($\hat{s}_y \to 1$), the module infers that high frequencies contain valid motion dynamics and amplifies them; for static actions ($\hat{s}_y \to 0$), it suppresses the gain to prevent artifact amplification.
The design of $\mathbf{G}$ is thus grounded in the frequency-selective filtering principle: rather than applying a uniform gain that cannot distinguish signal from noise in the high-frequency band, the semantic prior $\hat{s}_y$ provides a class-conditional switch that resolves this ambiguity in a theoretically principled manner.

\vspace{1mm}
\noindent \textbf{Internalizing Priors via Distillation.}
A critical challenge is obtaining the intensity score $\hat{s}_y$ during inference without relying on external large models. To achieve this, we \textbf{internalize} the LLM's commonsense knowledge about motion dynamics directly into the text encoder via knowledge distillation~\cite{zhang2022distilling,gou2021knowledge}.

We design a lightweight \textbf{Kinematic Projection Head} $\phi(\cdot)$ (a two-layer MLP) that maps the semantic text embedding $\mathbf{d}_y$ to a probability score $\hat{s}_y = \sigma(\phi(\mathbf{d}_y)) \in [0, 1]$, where $\sigma$ is the sigmoid function.
To supervise this head, we employ a simplified \textit{offline binary classification strategy}. Prior to training, we prompt an LLM to leverage its open-world knowledge to determine whether each action class involves significant high-frequency dynamics (assigning $s_y^{\text{GT}}=1$, e.g., {Punching/Slapping}) or remains structurally static (assigning $s_y^{\text{GT}}=0$, e.g., {Reading}).
During training, we minimize the binary cross-entropy loss between the predicted probability $\hat{s}_y$ and the binary label $s_y^{\text{GT}}$. 
This process effectively guides the text encoder to decode {implicit kinematic attributes} from sparse semantic labels. Consequently, during inference, we utilize the predicted probability directly as a continuous gain coefficient. 
It is worth noting that the core operations of SG-SRM—DCT/IDCT and element-wise modulation—are computationally efficient linear transformations. When implemented via optimized matrix multiplication, they introduce negligible latency compared to the backbone's attention mechanisms, ensuring the framework remains lightweight and scalable.

\subsection{Timestep-Adaptive Spectral Loss} \label{sec:loss}

While the architectural module enhances the model's capacity to represent high frequencies, the \emph{optimization objective} remains a critical bottleneck. Standard diffusion training relies on a uniform Mean Squared Error (MSE) loss. While theoretically capable of capturing all frequencies, in practice, MSE objectives suffer from the well-known {``regression to the mean''} problem in generative modeling~\cite{ zhang2018unreasonable}.
Since high-frequency micro-dynamics (e.g., rapid jitter or transient edges) are inherently stochastic, minimizing MSE drives the model to predict the \textit{statistical average} of all plausible variations. This averaging effect cancels out high-frequency details, resulting in oversmoothed predictions that preserve global pose structure but lack discriminative texture. Furthermore, neural networks exhibit a \textit{spectral bias}~\cite{rahaman2019spectral}, converging significantly faster on low-frequency components while struggling to optimize high-frequency errors under a uniform loss.

To counteract this, explicitly supervising the model in the frequency domain is necessary. However, a static frequency loss is suboptimal due to the intrinsic \textbf{coarse-to-fine} generative trajectory of diffusion models~\cite{choi2022perception}.
At large timesteps $t$ (high noise levels), the signal is dominated by noise, and the model focuses on establishing the global pose topology (low-frequency). Enforcing high-frequency consistency at this stage is counterproductive, as it forces the model to overfit to Gaussian noise. Conversely, at small timesteps $t$ (low noise levels), the structural foundation is established, and the model focuses on refining fine-grained details. A uniform frequency weight fails to respect this dynamic evolution.

We therefore propose a \textbf{Timestep-Adaptive Spectral Loss} that aligns the optimization focus with the denoising schedule. Instead of supervising the noise vector $\epsilon$, we impose constraints directly on the \textit{estimated clean signal} $\hat{\mathbf{z}}_0$. Following standard diffusion formulations~\cite{song2020ddim}, $\hat{\mathbf{z}}_0$ can be analytically recovered from the noisy latent $\mathbf{z}_t$ and the predicted noise $\epsilon_{\theta}(\mathbf{z}_t, t)$ via:
\begin{equation}\label{eq:pred_z0}
    \hat{\mathbf{z}}_0 = \frac{\mathbf{z}_t - \sqrt{1 - \bar{\alpha}_t} \epsilon_{\theta}(\mathbf{z}_t, t)}{\sqrt{\bar{\alpha}}_t}
\end{equation}
where $\bar{\alpha}_t$ denotes the cumulative noise schedule at timestep $t$. This formulation allows us to enforce spectral validity directly in the motion domain. The objective is formulated as:
\begin{equation}\label{eq:spectral_loss}
    \mathcal{L}_{\text{freq}} = \mathbb{E}_{t, \mathbf{z}_0} \left[ \sum_{k=0}^{L-1} \mathbf{W}(k, t) \cdot \| \text{DCT}(\mathbf{z}_0)_k - \text{DCT}(\hat{\mathbf{z}}_0)_k \|^2 \right]
\end{equation}
where $\mathbf{z}_0$ is the ground truth latent. The dynamic weighting mask $\mathbf{W}(k, t)$ is designed to progressively ``unlock'' high-frequency supervision:
\begin{equation}\label{eq:adaptive_weight}
    \mathbf{W}(k, t) = 
    \begin{cases} 
        1 & \text{if } k < M \quad (\text{Base Structural Loss}) \\
        \gamma \cdot (1 - \frac{t}{T}) & \text{if } k \ge M \quad (\text{Adaptive Detail Loss})
    \end{cases}
\end{equation}

This adaptive weighting is grounded in the observation that diffusion models follow a coarse-to-fine trajectory, where the reverse process transitions from global structure formation to local detail refinement~\cite{choi2022perception}. By linearly modulating the high-frequency loss weight according to the noise level, we suppress premature high-frequency hallucination during early stages while enforcing rigorous detail reconstruction as the generation approaches the clean data manifold.

Here, low frequencies ($k < M$) are consistently supervised to ensure structural stability. For high frequencies ($k \ge M$), the weight is modulated by a linear decay term $(1 - t/T)$. 
During the early noise-dominated stages ($t \to T$), the weight approaches zero, preventing the model from overfitting to noise. As the generation progresses ($t \to 0$), the weight gradually increases, penalizing the lack of sharp micro-dynamics in the final refinement stages.

\subsection{Curriculum-based Semantic Abstraction} \label{sec:curriculum}
While the proposed architectural and optimization mechanisms provide the \emph{spectral capacity} to synthesize high-frequency dynamics, a critical \textit{cognitive gap} remains. Zero-shot inference typically relies on sparse class names (e.g., ``Jump''), which are semantically {underspecified}. The same label may correspond to varying spectral signatures (e.g., tempo, intensity, rhythm) that are implicit in the name. Without explicit guidance, conditioning a diffusion generator solely on sparse labels encourages the denoiser to collapse towards a low-frequency ``average'' motion—semantically plausible but discriminatively weak.

To bridge this gap, we introduce a \textbf{Curriculum-based Semantic Abstraction} strategy. The core intuition is to provide the model with a \textit{semantic scaffold} during early training—using rich, kinematically explicit descriptions—and then progressively withdraw this guidance. This forces the visual encoder to {internalize} the correlation between sparse labels and complex motion priors, ensuring robustness when only sparse labels are available at test time.
\begin{figure}[t]
    \centering
    \includegraphics[width=0.9\linewidth]{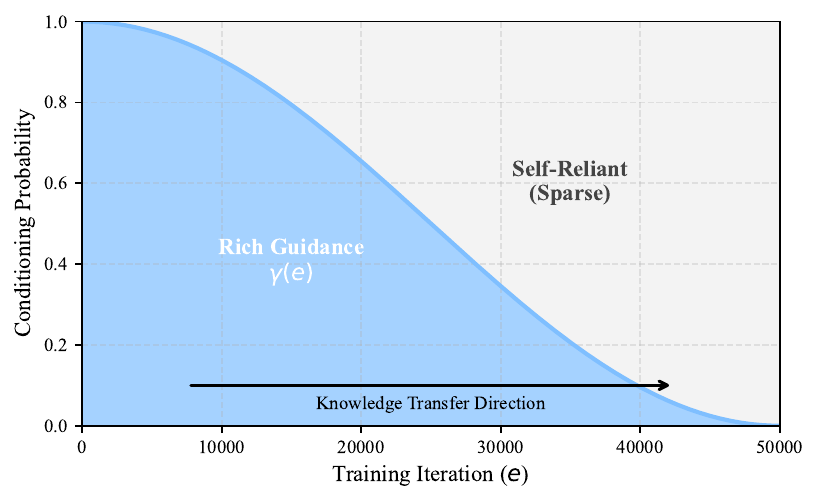}
    \caption{Illustration of the Curriculum-based Semantic Abstraction schedule. The probability $\gamma(e)$ of using rich descriptions follows a cosine annealing curve, starting high to provide a semantic scaffold and decaying to zero to enforce self-reliant zero-shot inference.}
    \label{fig:curriculum_curve}
\end{figure}

Formally, for each action class $y$, we construct two views: a sparse label prompt $\ell_y$ (the class name) and a set of LLM-generated rich descriptions $\mathcal{R}_y = \{r_y^{(n)}\}_{n=1}^{N_{desc}}$. These descriptions explicitly articulate kinematic details such as involved body parts, relative speed, and rhythm. The exact prompt templates used for LLM interaction are provided in Section~\ref{app:prompts}.
During training, we sample the conditioning text $c$ based on a dynamic probability schedule. We adopt the {cosine annealing schedule}~\cite{loshchilov2016sgdr} to ensure a smooth transition from teacher-guidance to self-reliance. Let $\gamma(e)$ be the probability of using rich descriptions at iteration $e$:
\begin{equation}\label{eq:curriculum_schedule}
    \gamma(e) = \frac{1}{2} \left( 1 + \cos\left( \frac{e \pi}{E_{\text{total}}} \right) \right)
\end{equation}
where $E_{\text{total}}$ is the total training iterations. The conditioning text is sampled as:
\begin{equation}\label{eq:curriculum_sampling}
    c = 
    \begin{cases} 
        r \sim \mathcal{R}_y & \text{with probability } \gamma(e) \quad (\text{Rich Guidance}) \\
        \ell_y & \text{with probability } 1 - \gamma(e) \quad (\text{Sparse Abstraction})
    \end{cases}
\end{equation}

As can be seen in \cref{fig:curriculum_curve}, in the initial phase ($\gamma \to 1$), the model learns to map visual features to explicit kinematic descriptions. As training progresses ($\gamma \to 0$), the model is forced to infer these dense kinematic features solely from the sparse token $\ell_y$, effectively transferring the rich semantic priors into the model's weights.
This stochastic switching instantiates the curriculum learning framework of Bengio~et~al.~\cite{bengio2009curriculum}, where training begins with richer supervisory signals and progressively transitions to sparser, more abstract conditions. The cosine annealing schedule~\cite{loshchilov2016sgdr} ensures a smooth, monotonic transition---analogous to progressive teacher-forcing reduction in sequence modeling---and guarantees that by the end of training the model operates solely under the zero-shot inference condition ($\gamma \to 0$).

\subsection{Training and Inference}
\label{sec:training}

\vspace{1mm}
\subsubsection{Two-Stage Optimization Strategy}
To ensure optimization stability and decouple semantic alignment from generative modeling, we adopt a two-stage training strategy.

\vspace{1mm}
\noindent \textbf{Stage 1: LLM's Prior Distillation.}
First, we train the lightweight projection head $\phi$ (\cref{sec:module}) to distill the LLM's commonsense kinematic knowledge. This is a fast, one-time pre-training step optimized via the binary cross-entropy loss: $\mathcal{L}_{\text{distill}} = - \left[ s_y^{\text{GT}} \log(\hat{s}_y) + (1 - s_y^{\text{GT}}) \log(1 - \hat{s}_y) \right]$, where $\hat{s}_y = \phi(\mathbf{d}_y)$. Once converged, $\phi$ is frozen and capable of predicting LLM's priors from text embeddings.

\vspace{1mm}
\noindent \textbf{Stage 2: Unified Generative Training.}
In the second stage, we train the main diffusion framework.
At each training step, we sample a batch of skeletal motion sequences $\mathbf{z}_0$ and their corresponding action classes $y$. 
To enforce the internalization of semantic priors, we determine the conditioning text $c$ based on the cosine annealing probability $\gamma(e)$ (\cref{sec:curriculum}).

\vspace{1mm}
\noindent\textbf{Forward \& Prediction:} The latent feature $\mathbf{z}_0$ is corrupted to $\mathbf{z}_t$ via the diffusion forward process. The network $\epsilon_\theta$ integrates the SG-SRM (\cref{sec:module}) to predict the noise. Specifically, we deploy this module appended to the output of each attention layer.
The total objective function focuses solely on the generative quality:
\begin{equation}\label{eq:total_loss}
    \mathcal{L}_{\text{total}} = \mathcal{L}_{\text{diff}} + \lambda_{\text{freq}} \cdot \mathcal{L}_{\text{freq}}
\end{equation}
where $\lambda_{\text{freq}}$ is a hyperparameter balancing the two terms.
The first term is the standard noise-prediction loss defined as $\mathcal{L}_{\text{diff}} = \| \epsilon_\theta (\mathbf{z}_t, t; \mathbf{d}_y, \hat{s}_y) - \bm{\epsilon} \|_2$. 
The second term $\mathcal{L}_{\text{freq}}$ is our \textit{Timestep-Adaptive Spectral Loss} (\cref{sec:loss}).
By optimizing $\mathcal{L}_{\text{total}}$, the model simultaneously learns to denoise the latent space (via $\mathcal{L}_{\text{diff}}$) and to reconstruct fine-grained motion dynamics (via $\mathcal{L}_{\text{freq}}$).

\subsubsection{Zero-Shot Inference}
\label{sec:infer}

For zero-shot recognition, we strictly follow the discriminative inference protocol established in TDSM~\cite{do2025bridging}. The core idea is to repurpose the diffusion model as a classifier by evaluating the noise reconstruction error for each candidate label.

Specifically, given an unseen skeleton $\mathbf{X}^{u}$, we encode it into the latent space and perturb it with fixed Gaussian noise $\bm{\epsilon}_{\text{test}}$ at a fixed timestep $t_{\text{test}}$. In all experiments, we set $t_{\text{test}}=25$ based on validation trends (see Section~\ref{app:ttest_selection} for a detailed sensitivity analysis) and keep this value unchanged across NTU, PKU-MMD, and Kinetics benchmarks. For each candidate action label $y' \in \mathcal{Y}_u$, we extract its text embedding $\mathbf{d}_{y'}$ and predict its kinematic intensity score $\hat{s}_{y'}$ using our projection head.
The network then predicts the noise conditioned on these inputs:
\begin{equation}\label{eq:inference_predict}
    \hat{\bm{\epsilon}}_{y'} = \mathcal{T}_{\text{diff}}(\mathbf{z}_{x,t}^{u}, t_{\text{test}}; \mathbf{d}_{y'}, \hat{s}_{y'})
\end{equation}
The final classification is performed by selecting the label that minimizes the $\ell_2$ distance between the added noise and the predicted noise:
\begin{equation}\label{eq:inference_classify}
    {\hat{y}}^{u} = \arg\min_{y' \in \mathcal{Y}_u} \| \bm{\epsilon}_{\text{test}} - \hat{\bm{\epsilon}}_{y'} \|_{2}
\end{equation}
This one-step inference avoids computationally expensive iterative sampling while effectively measuring the alignment between the skeleton dynamics and the candidate semantic prompts.

\section{Results}\label{sec:results}

\subsection{Datasets}

\noindent\textbf{NTU RGB+D \cite{Ntu60}.} The NTU RGB+D dataset (NTU-60) is a widely recognized large-scale benchmark for human action recognition, containing 56,880 action samples across 60 categories. It provides multi-modal data, including 3D skeletons, depth maps, and RGB videos captured by Kinect sensors \cite{Kinetics}, serving as a standard for evaluating both single- and multi-view recognition models. We adopt the cross-subject (X-sub) protocol, where the 40 subjects are equally partitioned into training and testing sets. Specifically, $\mathcal{D}_{\text{train}}$ is constructed from the training set using seen labels, while $\mathcal{D}_{\text{test}}$ is derived from the test set with unseen labels to ensure a rigorous zero-shot evaluation.

\noindent\textbf{NTU RGB+D 120 \cite{Ntu120}.} The NTU RGB+D 120 dataset (NTU-120) expands upon NTU-60 by incorporating 60 additional action classes, totaling 120 categories and 114,480 video samples. For the cross-subject (X-sub) evaluation, the 106 subjects are equally divided into training and testing sets. Following the same protocol as NTU-60, we utilize the training set with seen labels to form $\mathcal{D}_{\text{train}}$ and the test set with unseen labels to establish $\mathcal{D}_{\text{test}}$, maintaining a consistent zero-shot setting.

\noindent\textbf{PKU-MMD \cite{PKUMMD}.} The PKU-MMD dataset is a large-scale, multi-modal benchmark for action recognition, providing 3D skeleton data alongside RGB+D recordings. It involves 66 subjects, with 57 designated for training and 9 for testing. Adopting the cross-subject protocol to evaluate our framework's generalization, we construct $\mathcal{D}_{\text{train}}$ from seen labels and $\mathcal{D}_{\text{test}}$ from unseen labels.

\noindent\textbf{Kinetics-skeleton 200~\cite{yan2018spatial} \& Kinetics-skeleton 400 \cite{Kinetics}.} The Kinetics dataset is a large-scale collection of YouTube video clips covering a diverse range of human actions. For our experiments, we utilize the skeleton-based version where 2D joint locations are estimated from the RGB streams via OpenPose~\cite{yan2018spatial}.
Kinetics-400 includes 400 action classes, while Kinetics-200 is a subset comprising 200 categories. These datasets are significantly more challenging than NTU or PKU-MMD due to the unconstrained nature of the video capture and the potential noise in the estimated skeletons, providing a rigorous test for zero-shot generalization.

\begin{table*}[tbp]
    \Large
    \centering
    \resizebox{\textwidth}{!}{
    \def\arraystretch{1.2}
    \begin{tabular} {l|cc|cc|cc|cc}
        \hline
        \multirow{2}{*}{Methods} &\multicolumn{4}{c|}{NTU-60\;(Acc, \%)} & \multicolumn{4}{c}{NTU-120 \;(Acc, \%)} \\
        \cmidrule{2-9}
        & 55/5\phantom{$_{\pm 0.00}$} & 48/12\phantom{$_{\pm 0.00}$} & 40/20\phantom{$_{\pm 0.00}$} & 30/30\phantom{$_{\pm 0.00}$} & 110/10\phantom{$_{\pm 0.00}$} & 96/24\phantom{$_{\pm 0.00}$} & 80/40\phantom{$_{\pm 0.00}$} & 60/60\phantom{$_{\pm 0.00}$} \\
        \hline
        ReViSE \cite{ReViSE} & 53.91\phantom{$_{\pm 0.00}$} & 17.49\phantom{$_{\pm 0.00}$} & 24.26\phantom{$_{\pm 0.00}$} & 14.81\phantom{$_{\pm 0.00}$} & 55.04\phantom{$_{\pm 0.00}$} & 32.38\phantom{$_{\pm 0.00}$} & 19.47\phantom{$_{\pm 0.00}$} & 8.27\phantom{$_{\pm 0.00}$} \\
        JPoSE \cite{JPoSE} & 64.82\phantom{$_{\pm 0.00}$} & 28.75\phantom{$_{\pm 0.00}$} & 20.05\phantom{$_{\pm 0.00}$} & 12.39\phantom{$_{\pm 0.00}$} & 51.93\phantom{$_{\pm 0.00}$} & 32.44\phantom{$_{\pm 0.00}$} & 13.71\phantom{$_{\pm 0.00}$} & 7.65\phantom{$_{\pm 0.00}$} \\
        CADA-VAE \cite{CADAVAE} & 76.84\phantom{$_{\pm 0.00}$} & 28.96\phantom{$_{\pm 0.00}$} & 16.21\phantom{$_{\pm 0.00}$} & 11.51\phantom{$_{\pm 0.00}$} & 59.53\phantom{$_{\pm 0.00}$} & 35.77\phantom{$_{\pm 0.00}$} & 10.55\phantom{$_{\pm 0.00}$} & 5.67\phantom{$_{\pm 0.00}$} \\
        SynSE \cite{SynSE} & 75.81\phantom{$_{\pm 0.00}$} & 33.30\phantom{$_{\pm 0.00}$} & 19.85\phantom{$_{\pm 0.00}$} & 12.00\phantom{$_{\pm 0.00}$} & 62.69\phantom{$_{\pm 0.00}$} & 38.70\phantom{$_{\pm 0.00}$} & 13.64\phantom{$_{\pm 0.00}$} & 7.73\phantom{$_{\pm 0.00}$} \\
        SMIE \cite{SMIE} & 77.98\phantom{$_{\pm 0.00}$} & 40.18\phantom{$_{\pm 0.00}$} & -\phantom{$_{\pm 0.00}$} & -\phantom{$_{\pm 0.00}$} & 65.74\phantom{$_{\pm 0.00}$} & 45.30\phantom{$_{\pm 0.00}$} & -\phantom{$_{\pm 0.00}$} & -\phantom{$_{\pm 0.00}$} \\
        PURLS \cite{PURLS} & 79.23\phantom{$_{\pm 0.00}$} & 40.99\phantom{$_{\pm 0.00}$} & 31.05\phantom{$_{\pm 0.00}$} & 23.52\phantom{$_{\pm 0.00}$} & 71.95\phantom{$_{\pm 0.00}$} & 52.01\phantom{$_{\pm 0.00}$} & 28.38\phantom{$_{\pm 0.00}$} & 19.63\phantom{$_{\pm 0.00}$} \\
        SA-DVAE \cite{SADVAE} & 82.37\phantom{$_{\pm 0.00}$} & 41.38\phantom{$_{\pm 0.00}$} & -\phantom{$_{\pm 0.00}$} & -\phantom{$_{\pm 0.00}$} & 68.77\phantom{$_{\pm 0.00}$} & 46.12\phantom{$_{\pm 0.00}$} & -\phantom{$_{\pm 0.00}$} & -\phantom{$_{\pm 0.00}$} \\
        STAR \cite{STAR} & 81.40\phantom{$_{\pm 0.00}$} & 45.10\phantom{$_{\pm 0.00}$} & -\phantom{$_{\pm 0.00}$} & -\phantom{$_{\pm 0.00}$} & 63.30\phantom{$_{\pm 0.00}$} & 44.30\phantom{$_{\pm 0.00}$} & -\phantom{$_{\pm 0.00}$} & -\phantom{$_{\pm 0.00}$} \\
        DVTA \cite{DVTA} & 79.28\phantom{$_{\pm 0.00}$} & 44.14\phantom{$_{\pm 0.00}$} & 32.68\phantom{$_{\pm 0.00}$} & 24.16\phantom{$_{\pm 0.00}$} & \underline{74.89}\phantom{$_{\pm 0.00}$} & 51.81\phantom{$_{\pm 0.00}$} & 28.89\phantom{$_{\pm 0.00}$} & 18.43\phantom{$_{\pm 0.00}$} \\
        InfoCPL \cite{InfoCPL} & 85.91\phantom{$_{\pm 0.00}$} & 53.32\phantom{$_{\pm 0.00}$} & 36.03\phantom{$_{\pm 0.00}$} & 25.44\phantom{$_{\pm 0.00}$} & 74.81\phantom{$_{\pm 0.00}$} & 60.05\phantom{$_{\pm 0.00}$} & 36.81\phantom{$_{\pm 0.00}$} & 24.72\phantom{$_{\pm 0.00}$} \\
        FS-VAE \cite{wu2025frequency} & \underline{86.90}\phantom{$_{\pm 0.00}$} & \underline{57.20}\phantom{$_{\pm 0.00}$} & \underline{36.17}\phantom{$_{\pm 0.00}$} & 25.72\phantom{$_{\pm 0.00}$} & 74.40\phantom{$_{\pm 0.00}$} & 62.50\phantom{$_{\pm 0.00}$} & \underline{37.06}\phantom{$_{\pm 0.00}$} & 26.45\phantom{$_{\pm 0.00}$} \\
        TDSM~\cite{do2025bridging} & {86.49}\phantom{$_{\pm 0.00}$} & {56.03}\phantom{$_{\pm 0.00}$} & {36.09}\phantom{$_{\pm 0.00}$} & {\underline{25.88}}\phantom{$_{\pm 0.00}$} & {74.15}\phantom{$_{\pm 0.00}$} & {\underline{65.06}}\phantom{$_{\pm 0.00}$} & {36.95}\phantom{$_{\pm 0.00}$} & {\underline{27.21}}\phantom{$_{\pm 0.00}$} \\
        \textbf{FDSM(Ours)} & \textbf{87.79}$_{\pm 0.08}$ & \textbf{57.46}$_{\pm 0.07}$ & \textbf{37.42}$_{\pm 0.08}$ & \textbf{26.55}$_{\pm 0.09}$ & \textbf{75.24}$_{\pm 0.07}$ & \textbf{66.52}$_{\pm 0.06}$ & \textbf{39.16}$_{\pm 0.10}$ & \textbf{28.67}$_{\pm 0.08}$ \\
        \hline
    \end{tabular} }
    \caption{Top-1 accuracy results of various zero-shot skeleton-based action recognition (ZSAR) methods evaluated on the SynSE and PURLS benchmarks for the NTU-60 and NTU-120 datasets. Each split is denoted as X/Y, where X represents the number of seen classes and Y the number of unseen classes. The results in \textbf{bold} highlight the best-performing model. For our method, the reported accuracy is the average value obtained from 10 trials, each with different Gaussian noise.}
  \label{tab:main}
\end{table*}

\subsection{Experiment Setup}
FDSM was implemented in PyTorch \cite{Pytorch} and evaluated using a single NVIDIA L40s (48GB) GPU. All model variants were trained for 50,000 iterations, including a 100-step warm-up phase. We utilized the AdamW optimizer \cite{AdamW} with a learning rate of $1 \times 10^{-4}$ and a weight decay of 0.01, modulated by a cosine-annealing scheduler \cite{CosineAnneal}. The training batch size was 256. Following~\cite{do2025bridging}, the diffusion process was trained with $T=50$ timesteps, while the inference timestep was fixed to $t_{\text{test}} = 25$ for all datasets.
Following TDSM~\cite{do2025bridging}, the backbone is a 12-layer Diffusion Transformer with a latent dimension of 768. We append our SG-SRM module after each attention layer in the backbone, setting the frequency split index $M=L/4$ (\cref{eq:gain_filter}), the spectral residual gain factor $\alpha=1.0$ (\cref{eq:spectral_gain}), the timestep-adaptive weight coefficient $\gamma=1.0$ (\cref{eq:adaptive_weight}), and the spectral loss weight $\lambda_{\text{freq}}=1.0$ (\cref{eq:total_loss}). For the Curriculum-based Semantic Abstraction (\cref{eq:curriculum_schedule}), we set the total training iterations $E_{\text{total}}=50{,}000$ to match our training schedule, and generated rich descriptions per action class using GPT-4 \cite{achiam2023gpt}.
To reduce stochastic variance at inference, each reported result is averaged over 10 Gaussian noise initializations.

\noindent\textbf{Kinematic Projection Head Setup.} The projection head $\phi$ consists of a 2-layer MLP with a hidden dimension of 256 and a ReLU activation, followed by a sigmoid layer. To train this head, we curated a binary labeled dataset by prompting GPT-4 to categorize all action classes in NTU-60/120 and PKU-MMD based on their motion intensity (1 for dynamic/jittery, 0 for static). A detailed breakdown of the LLM-derived intensity labels across all benchmarks is provided in Section~\ref{app:intensity_stats}. The head was pre-trained for 500 epochs using the binary cross-entropy loss with a learning rate of $1 \times 10^{-3}$, and then frozen during the main diffusion training phase.

To ensure a fair comparison with existing zero-shot action recognition (ZSAR) methods, we adopted Shift-GCN \cite{ShiftGCN} and ST-GCN \cite{STGCN} as skeleton encoders for the SynSE \cite{SynSE}/PURLS \cite{PURLS} and SMIE \cite{SMIE} settings, respectively. Furthermore, we utilized the same text prompts and text encoder from CLIP \cite{CLIP, openclip} as prior works to maintain consistency in semantic representations. Unless otherwise specified, the best and second-best results in all tables are highlighted in bold and underlined, respectively.

\begin{table}[tbp]
    \centering
    \def\arraystretch{1.0}
    \begin{tabular} {l|c|c|c}
        \hline
        \multirow{2}{*}{Methods} & NTU-60\;(Acc, \%) & NTU-120\;(Acc, \%) & PKU-MMD\;(Acc, \%) \\
        \cmidrule{2-4}
        & 55/5 split\phantom{$_{\pm 0.00}$} & 110/10 split\phantom{$_{\pm 0.00}$} & 46/5 split\phantom{$_{\pm 0.00}$} \\
        \hline
        ReViSE \cite{ReViSE} & 60.94\phantom{$_{\pm 0.00}$} & 44.90\phantom{$_{\pm 0.00}$} & 59.34\phantom{$_{\pm 0.00}$} \\
        JPoSE \cite{JPoSE} & 59.44\phantom{$_{\pm 0.00}$} & 46.69\phantom{$_{\pm 0.00}$} & 57.17\phantom{$_{\pm 0.00}$} \\
        CADA-VAE \cite{CADAVAE} & 61.84\phantom{$_{\pm 0.00}$} & 45.15\phantom{$_{\pm 0.00}$} & 60.74\phantom{$_{\pm 0.00}$} \\
        SynSE \cite{SynSE} & 64.19\phantom{$_{\pm 0.00}$} & 47.28\phantom{$_{\pm 0.00}$} & 53.85\phantom{$_{\pm 0.00}$} \\
        SMIE \cite{SMIE} & 65.08\phantom{$_{\pm 0.00}$} & 46.40\phantom{$_{\pm 0.00}$} & 60.83\phantom{$_{\pm 0.00}$} \\
        SA-DVAE \cite{SADVAE} & 84.20\phantom{$_{\pm 0.00}$} & 50.67\phantom{$_{\pm 0.00}$} & 66.54\phantom{$_{\pm 0.00}$} \\
        STAR \cite{STAR} & 77.50\phantom{$_{\pm 0.00}$} & -\phantom{$_{\pm 0.00}$} & 70.60\phantom{$_{\pm 0.00}$} \\
        DVTA  \cite{DVTA} & 74.03\phantom{$_{\pm 0.00}$} & 60.33\phantom{$_{\pm 0.00}$} & \underline{77.06}\phantom{$_{\pm 0.00}$}\\
        InfoCPL \cite{InfoCPL} & 80.96\phantom{$_{\pm 0.00}$} & \underline{70.07}\phantom{$_{\pm 0.00}$} & \textbf{85.15}\phantom{$_{\pm 0.00}$}\\
        FS-VAE \cite{wu2025frequency} & 87.63\phantom{$_{\pm 0.00}$} & 69.72\phantom{$_{\pm 0.00}$} & 70.97\phantom{$_{\pm 0.00}$}\\
        {TDSM}~\cite{do2025bridging} & {\underline{88.88}}\phantom{$_{\pm 0.00}$} & {69.47}\phantom{$_{\pm 0.00}$} & {70.76}\phantom{$_{\pm 0.00}$} \\
        \textbf{FDSM(Ours)} & \textbf{90.13}$_{\pm 0.07}$ & \textbf{70.59}$_{\pm 0.08}$ & 72.18$_{\pm 0.12}$ \\
        \hline
    \end{tabular}
    \caption{Top-1 accuracy results of various ZSAR methods evaluated on the NTU-60, NTU-120, and PKU-MMD datasets under the SMIE benchmark. The reported values are the average performance across three splits.}
  \label{tab:smie}
\end{table}

\subsection{Performance Evaluation}

\noindent\textbf{Evaluation on SynSE \cite{SynSE} and PURLS \cite{PURLS} benchmarks.} Table~\ref{tab:main} presents a comprehensive comparison between FDSM and state-of-the-art ZSAR methods on the SynSE and PURLS benchmark protocols. 
The SynSE protocol comprises standard seen/unseen ratios: 55/5 and 48/12 on NTU-60, and 110/10 and 96/24 on NTU-120. In contrast, the PURLS protocol presents more extreme scenarios: 40/20 and 30/30 on NTU-60, and 80/40 and 60/60 on NTU-120. 
As demonstrated in Table~\ref{tab:main}, FDSM consistently outperforms the SOTA, TDSM, across all settings. 
This performance gap stems from the different approaches to visual-semantic alignment. TDSM~\cite{do2025bridging} uses a discriminative triplet diffusion loss ($L_{TD}$) to enforce class separation, but it does not explicitly address the ``low-pass'' spectral bias of standard Diffusion Transformers. This often leads to oversmoothed motion representations, losing high-frequency details like the wrist movements in ``\textit{giving something to other person}'' or leg jitters in ``\textit{kicking}''. FDSM, on the other hand, employs SG-SRM to recover these missing spectral bands. Our results indicate that for zero-shot generalization, reconstructing fine-grained motion details is more effective than focusing solely on discriminative margins, particularly when unseen actions are distinguished by micro-dynamics rather than global pose.

\begin{figure}[t]
    \centering
    \safeincludegraphics[width=0.9\linewidth]{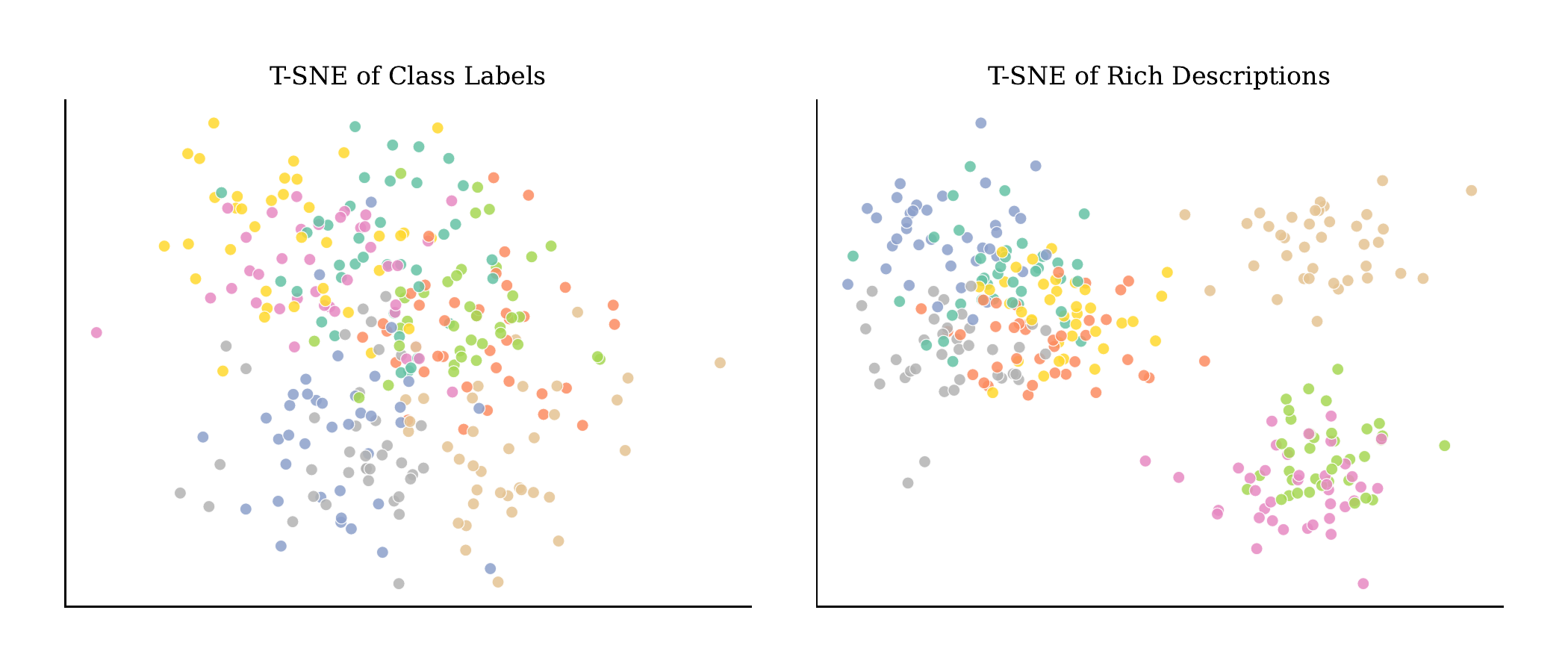}
    \caption{T-SNE visualization of textual embeddings on NTU RGB+D 120. LLM-enriched descriptions (Right) form better-separated clusters compared to simple class labels (Left), facilitating more robust cross-modal alignment.}
    \label{fig:tsne}
\end{figure}

To further validate the effectiveness of our Curriculum-based Semantic Abstraction, Fig.~\ref{fig:tsne} presents T-SNE visualizations comparing CLIP embeddings of simple class labels versus LLM-enriched rich descriptions on NTU RGB+D 120. The enriched descriptions form better-separated intra-class clusters, confirming that richer kinematic semantics improve the discriminability of the textual embedding space and provide stronger semantic guidance for spectral-aware generation.

\noindent\textbf{Evaluation on SMIE \cite{SMIE} benchmark.} The SMIE benchmark serves as a rigorous test for the stability of ZSAR models across varied unseen label distributions. As shown in Table~\ref{tab:smie}, FDSM consistently outperforms state-of-the-art competitors, including TDSM, across NTU-60, NTU-120, and PKU-MMD. Specifically, FDSM achieves a 1.42\% improvement on PKU-MMD, a dataset characterized by shorter sequences and more compact motion. Our method's superior performance across these diverse splits demonstrates that explicit frequency modulation provides a more universal motion representation than standard contrastive alignment, allowing the model to adapt to different skeletal sampling rates and temporal scales inherent in different datasets.

\begin{table}[tbp]
    \scriptsize
    \centering
    \def\arraystretch{1.0}
    \begin{tabular} {l|c|c|c|c}
        \Xhline{2\arrayrulewidth}
        \multirow{2}{*}{Methods} &\multicolumn{4}{c}{Kinetics-200\;(Acc, \%)} \\
        \cmidrule{2-5}
        & 180/20 split & 160/40 split & 140/60 split & 120/80 split \\
        \hline
        ReViSE \cite{ReViSE} & 24.95 & 13.28 & 8.14 & 6.23 \\
        DeViSE \cite{DeViSE} & 22.22 & 12.32 & 7.97 & 5.65 \\
        PURLS \cite{PURLS} & 32.22 & 22.56 & 12.01 & 11.75 \\
        TDSM~\cite{do2025bridging} & {{38.18}} & {{24.43}} & {{15.28}} & 13.09 \\
        \textbf{FDSM(Ours)} & \textbf{39.76} & \textbf{27.01} & \textbf{16.96} & \textbf{14.73} \\
        \Xhline{2\arrayrulewidth}
    \end{tabular}
    \caption{Top-1 accuracy results of FDSM evaluated on the Kinetics-200 dataset under the PURLS \cite{PURLS} benchmark.}
    \label{tab:kinetics1}
\end{table}

\begin{table}[tbp]
    \scriptsize
    \centering
    \def\arraystretch{1.1}
    \begin{tabular} {l|c|c|c|c}
        \Xhline{2\arrayrulewidth}
        \multirow{2}{*}{Methods} &\multicolumn{4}{c}{Kinetics-400\;(Acc, \%)} \\
        \cmidrule{2-5}
        & 360/40 split & 320/80 split & 300/100 split & 280/120 split \\
        \hline
        ReViSE \cite{ReViSE} & 20.84 & 11.82 & 9.49 & 8.23 \\
        DeViSE \cite{DeViSE} & 18.37 & 10.23 & 9.47 & 8.34 \\
        PURLS \cite{PURLS}  & 34.51 & 24.32 & 16.99 & 14.28 \\
        TDSM~\cite{do2025bridging}  & 38.92 & 26.24 & 18.45 & 16.10 \\
        \textbf{FDSM(Ours)}  & \textbf{40.25} & \textbf{28.17} & \textbf{20.13} & \textbf{17.88} \\
        \Xhline{2\arrayrulewidth}
    \end{tabular}
    \caption{Top-1 accuracy results of FDSM evaluated on the Kinetics-400 dataset under the PURLS \cite{PURLS} benchmark.}
    \label{tab:kinetics2}
\end{table}

\noindent\textbf{Evaluation on Large-scale Kinetics-skeleton Benchmarks.} 
To further evaluate the robustness of FDSM in unconstrained environments, we report results on Kinetics-200 and Kinetics-400 in Table~\ref{tab:kinetics1} and Table~\ref{tab:kinetics2}. Unlike the laboratory-captured NTU datasets, Kinetics features significant background clutter and severe pose estimation inaccuracies. Despite these challenges, FDSM sets a new state-of-the-art across all splits. For instance, on the Kinetics-400 360/40 split, FDSM achieves 40.25\% accuracy, surpassing TDSM by 1.33\%. This robustness is primarily attributed to our Timestep-Adaptive Spectral Loss. By dynamically suppressing high-frequency gradients during the early stages of denoising, the model avoids overfitting to the high-frequency artifacts and sensor jitter common in estimated skeletons, while still recovering essential motion textures in the final refinement steps. This ensures that the synthesized features remain semantically pure yet kinematically rich.

\begin{table}[tbp]
    \scriptsize
    \centering
    \def\arraystretch{1.2}
    \begin{tabular} {l|c|c|c|c|c|c}
        \hline
        \multirow{2}{*}{Methods} &\multicolumn{3}{c|}{NTU-60\;(Acc, \%)} & \multicolumn{3}{c}{NTU-120 \;(Acc, \%)} \\
        \cmidrule{2-7}
        & 55/5 & 48/12 & 30/30 & 110/10 & 96/24 & 60/60 \\
        \hline
        Full model & \textbf{87.79} & \textbf{57.46} & \textbf{26.55} & \textbf{75.24} & \textbf{66.52} & \textbf{28.67} \\
        \hline
        w.o. SG-SRM & 86.52 & 56.11 & 25.42 & 74.18 & 65.13 & 27.53 \\
        w.o. $\mathcal{L}_{\text{Freq}}$ & 86.45 & 55.98 & 25.35 & 74.21 & 65.08 & 27.48 \\
        w.o. Curriculum & 87.31 & 57.18 & 26.11 & 74.89 & 66.15 & 28.32 \\
        \hline
    \end{tabular}
    \caption{Ablation study on the contribution of each component. SG-SRM: Semantic-Guided Spectral Residual Module; $\mathcal{L}_{\text{Freq}}$: Timestep-Adaptive Spectral Loss; Curriculum: Curriculum-based Semantic Abstraction. We report the performance degradation when removing each component from the full model.}
  \label{tab:loss}
\end{table}

\subsection{Ablation Studies and Analysis}

\begin{figure}[t]
    \centering
    \safeincludegraphics[width=\linewidth]{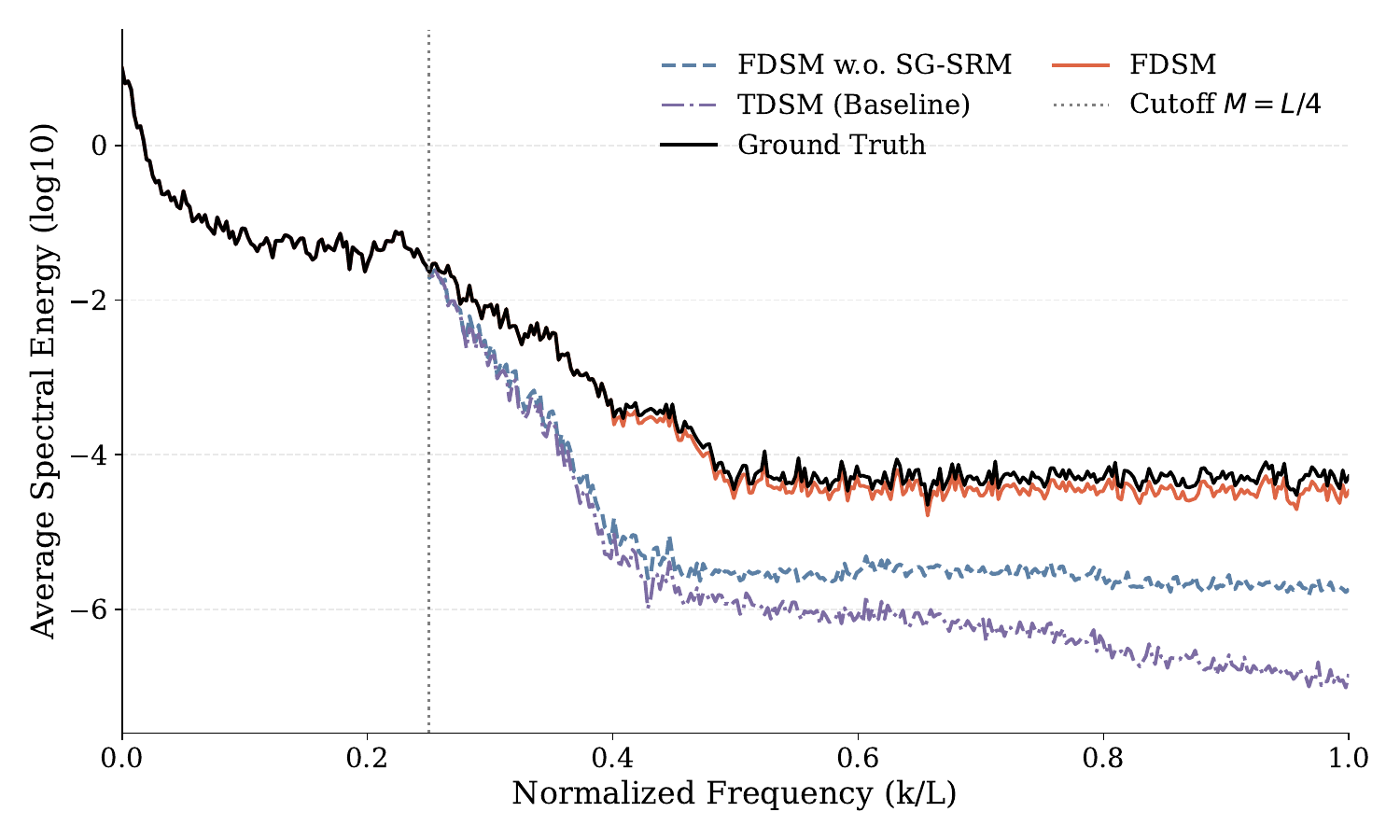}
    \caption{DCT-based spectral energy of ST-GCN latent features along the temporal dimension, where $k$ denotes the frequency index and $M{=}L/4$ is the cutoff separating low- and high-frequency bands. All spectral profiles track closely in the low normalized-frequency band ($k/L < 0.25$), confirming that all models capture global pose structure. Beyond the cutoff ($k/L \geq 0.25$), GT latents maintain substantial energy while both TDSM and FDSM without SG-SRM exhibit a sharp drop---this systematic gap is the \emph{Spectral Bias}. FDSM (full model) recovers the suppressed high-frequency energy, closely tracking the GT curve.}
    \label{fig:psd}
\end{figure}

\noindent\textbf{Contribution of each component.}
Table~\ref{tab:loss} isolates the impact of individual components. The results highlight that the \textbf{SG-SRM} and \textbf{$\mathcal{L}_{\text{Freq}}$} are the cornerstone contributions; removing either component precipitates a sharp performance drop, reverting metrics to near-baseline (TDSM) levels. This implies that the architectural capacity enhancement and the spectral optimization objective are mutually dependent---one provides the physical mechanism for high-frequency synthesis, while the other provides the necessary supervision. Fig.~\ref{fig:psd} provides direct spectral evidence for SG-SRM's role. Spectral decomposition refers to the decomposition of a temporal signal into orthogonal frequency components via DCT on ST-GCN latent features, separating global pose topology ($k < L/4$) from fine-grained micro-dynamics ($k \geq L/4$) such as joint velocity bursts. The \emph{natural motion distribution} refers to the frequency-domain energy profile of real human motion: since ground-truth (GT) sequences are captured directly from real movements, their per-frequency energy $|c_k^{\mathrm{GT}}|^2$ serves as the empirical oracle of how energy should be distributed across all frequency bands in realistic motion. As shown in Fig.~\ref{fig:psd}, all spectral profiles track closely in the low normalized-frequency band ($k/L < 0.25$), confirming that all models---including TDSM---successfully capture global pose structure. However, beyond the cutoff ($k/L \geq 0.25$), GT latents maintain substantial energy while both TDSM and FDSM without SG-SRM exhibit a sharp drop: this systematic gap is precisely the Spectral Bias. Critically, this deficit is invisible in the temporal domain---because time-domain representations intermix all frequency components, any high-frequency deficiency remains indistinguishable from the dominant low-frequency structure~\cite{rahaman2019spectral}. DCT-based spectral decomposition explicitly disentangles these components, making the per-band energy gap directly measurable at each frequency index $k$ via $|c_k|^2$, which is why spectral analysis is the appropriate diagnostic tool here.

\begin{figure}[h]
    \centering
    \includegraphics[width=0.85\textwidth]{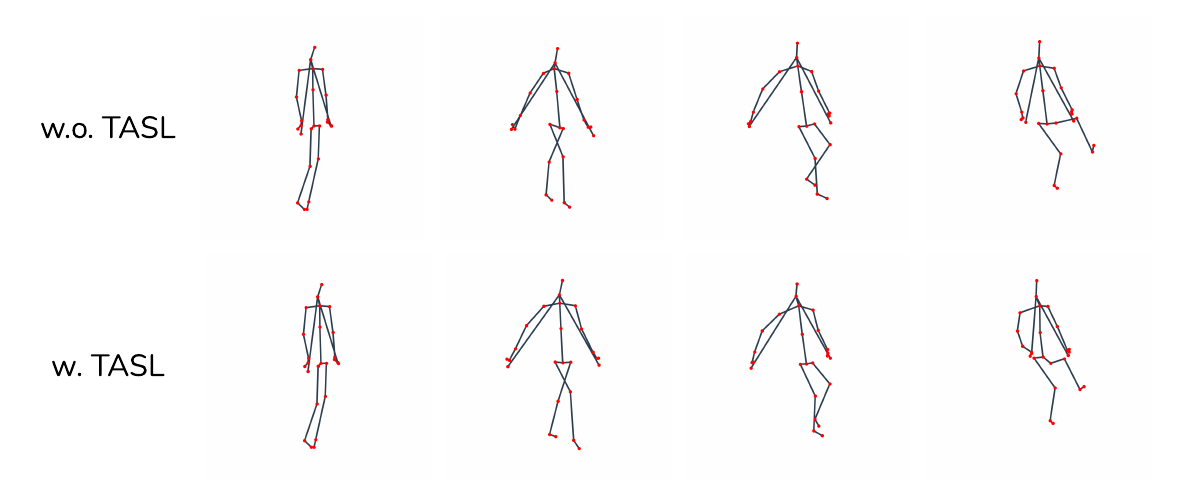}
    \caption{Qualitative comparison of decoded latent trajectories predicted by FDSM without (top) and with (bottom) the Timestep-Adaptive Spectral Loss ($\mathcal{L}_{\text{freq}}$) on a kicking sequence from NTU RGB+D 120. Without TASL, the predicted $\hat{z}_0$ trajectory collapses to an oversmoothed average, yielding nearly identical frames with minimal inter-frame variation. With TASL, the adaptive high-frequency supervision (Eq.~\ref{eq:adaptive_weight}) recovers richer micro-dynamics, including discriminative limb articulations that distinguish kinematically similar actions.}
    \label{fig:tasl_vis}
\end{figure}

Fig.~\ref{fig:tasl_vis} complements this spectral analysis with a qualitative view of $\mathcal{L}_{\text{freq}}$'s impact on the decoded motion trajectory. Without TASL (top), the model collapses toward an averaged pose: consecutive frames are nearly indistinguishable, with minimal arm swing and leg displacement. With TASL (bottom), the adaptive high-frequency supervision (Eq.~9) progressively enforces fine-grained detail as $t \to 0$, recovering discriminative micro-dynamics such as the kicking leg trajectory and arm counter-motion across frames.

In contrast, the \textbf{Curriculum-based Semantic Abstraction} yields a moderate but consistent improvement. This suggests that while semantic scaffolding refines the alignment, the core performance gains are primarily driven by our frequency-aware structural designs. 

\begin{table}[tbp]
    \scriptsize
    \centering
    \def\arraystretch{1.2}
    \begin{tabular} {c|c|c|c|c|c|c}
        \hline
        \multirow{2}{*}{\makecell{Gaussian\\noise $\bm\epsilon$}} & \multicolumn{3}{c|}{NTU-60\;(Acc, \%)} & \multicolumn{3}{c}{NTU-120 \;(Acc, \%)} \\
        \cmidrule{2-7}
        & 55/5 & 48/12 & 30/30 & 110/10 & 96/24 & 60/60 \\
        \hline
        Fixed & 77.68 & 45.52 & 19.82 & 65.18 & 53.47 & 18.55 \\
        Random(Ours) & \textbf{87.79} & \textbf{57.46} & \textbf{26.55} & \textbf{75.24} & \textbf{66.52} & \textbf{28.67} \\
        \hline
    \end{tabular}
    \caption{Ablation study on the effect of noise $\bm{\epsilon}$ during training.}
  \label{tab:fixed}
\end{table}

\begin{table}[tbp]
    \scriptsize
    \centering
    \def\arraystretch{1.2}
    \begin{tabular} {l|c|c|c|c|c|c}
        \hline
        \multirow{2}{*}{Cutoff $M$} & \multicolumn{3}{c|}{NTU-60\;(Acc, \%)} & \multicolumn{3}{c}{NTU-120 \;(Acc, \%)} \\
        \cmidrule{2-7}
        & 55/5 & 48/12 & 30/30 & 110/10 & 96/24 & 60/60 \\
        \hline
        $L/8$  & 87.45 & 57.10 & 26.25 & 74.95 & 66.20 & 28.40 \\
        \textbf{$L/4$(Ours) } & \textbf{87.79} & \textbf{57.46} & \textbf{26.55} & \textbf{75.24} & \textbf{66.52} & \textbf{28.67} \\
        $L/2$  & 87.10 & 56.85 & 25.95 & 74.60 & 65.90 & 28.05 \\
        \hline
    \end{tabular}
    \caption{Sensitivity to the frequency cutoff parameter $M$.}
  \label{tab:cutoff}
\end{table}

\noindent\textbf{Effect of random Gaussian noise.} To investigate the role of stochasticity during training, we performed an ablation study by utilizing fixed Gaussian noise instead of sampling new random noise at each iteration. As shown in Table~\ref{tab:fixed}, employing static noise patterns oversimplifies the optimization task, causing the network to memorize specific noise realizations and undermining its generalizability. Conversely, introducing random Gaussian noise at each step serves as an effective regularization mechanism by increasing data variability. This stochasticity prevents overfitting, enhances model robustness, and facilitates superior alignment between skeleton features and semantic text prompts.


\noindent\textbf{Sensitivity of Frequency Cutoff $M$.} 
Table~\ref{tab:cutoff} investigates the sensitivity of the frequency partition point $M$ in SG-SRM. $M$ defines the boundary between the ``Structural Base''(low frequency) and ``Adaptive Details'' (high frequency). We observe that $M=L/4$ achieves the optimal performance. When $M$ is too small (e.g., $L/8$), the module treats too many structural components as high-frequency noise, potentially leading to instability. Conversely, when $M$ is too large (e.g., $L/2$), many discriminative high-frequency details are mistakenly categorized as base structure and thus not amplified by SG-SRM, resulting in a loss of fine-grained information.

\noindent\textbf{Analysis of Curriculum Schedule.}
To validate the effectiveness of the proposed cosine annealing strategy for curriculum learning, we compared it against alternative schedules: (1) \textit{Linear Decay}, where the probability of using rich descriptions decreases linearly; (2) \textit{Step Decay}, where the probability drops by half at fixed intervals; and (3) \textit{Fixed Probability}, where the probability is held constant at 0.5. As shown in Table~\ref{tab:curriculum_schedule}, the cosine annealing schedule yields the best performance. This confirms that a smooth, non-linear transition from rich semantic scaffolding to sparse label reliance is optimal for internalizing kinematic priors without causing sudden shifts in the optimization landscape.

\begin{table}[tbp]
    \scriptsize
    \centering
    \def\arraystretch{1.2}
    \begin{tabular} {l|c|c|c|c|c|c}
        \hline
        \multirow{2}{*}{Schedule} & \multicolumn{3}{c|}{NTU-60\;(Acc, \%)} & \multicolumn{3}{c}{NTU-120 \;(Acc, \%)} \\
        \cmidrule{2-7}
        & 55/5 & 48/12 & 30/30 & 110/10 & 96/24 & 60/60 \\
        \hline
        Fixed (0.5) & 87.12 & 56.95 & 25.85 & 74.75 & 66.00 & 28.15 \\
        Step Decay & 87.35 & 57.10 & 26.15 & 74.90 & 66.15 & 28.35 \\
        Linear Decay & 87.55 & 57.25 & 26.35 & 75.05 & 66.30 & 28.50 \\
        \textbf{Cosine (Ours)} & \textbf{87.79} & \textbf{57.46} & \textbf{26.55} & \textbf{75.24} & \textbf{66.52} & \textbf{28.67} \\
        \hline
    \end{tabular}
    \caption{Ablation on curriculum schedules.}
  \label{tab:curriculum_schedule}
\end{table}

\noindent\textbf{Analysis of Semantic Gating Mechanism.}
To justify the design of our Semantic-Guided Spectral Residual Module (SG-SRM), we evaluated different gating strategies for the high-frequency gain $\mathbf{G}_k$: (1) \textit{Uniform Gain}, where $\mathbf{G}_k=1$ for all high frequencies (equivalent to unconditioned amplification); (2) \textit{Random Gain}, where $\mathbf{G}_k$ is sampled randomly; and (3) \textit{No Gain}, where $\mathbf{G}_k=0$ (equivalent to removing SG-SRM). Table~\ref{tab:gating} demonstrates that our semantic-aware gating significantly outperforms uniform or random strategies. This supports our hypothesis that high-frequency amplification must be selectively applied based on the kinematic context to avoid amplifying sensor noise in static actions.

\begin{table}[tbp]
    \scriptsize
    \centering
    \def\arraystretch{1.2}
    \begin{tabular} {l|c|c|c|c|c|c}
        \hline
        \multirow{2}{*}{Gating Strategy} & \multicolumn{3}{c|}{NTU-60\;(Acc, \%)} & \multicolumn{3}{c}{NTU-120 \;(Acc, \%)} \\
        \cmidrule{2-7}
        & 55/5 & 48/12 & 30/30 & 110/10 & 96/24 & 60/60 \\
        \hline
        No Gain ($s=0$) & 86.52 & 56.11 & 25.42 & 74.18 & 65.13 & 27.53 \\
        Random Gain & 84.50 & 53.80 & 23.50 & 72.10 & 63.50 & 25.80 \\
        Uniform Gain ($s=1$) & 85.80 & 55.10 & 24.85 & 73.40 & 64.80 & 26.90 \\
        \textbf{Predicted (Ours)} & \textbf{87.79} & \textbf{57.46} & \textbf{26.55} & \textbf{75.24} & \textbf{66.52} & \textbf{28.67} \\
        \hline
    \end{tabular}
    \caption{Ablation study on the semantic gating mechanism in SG-SRM.}
  \label{tab:gating}
\end{table}

\noindent\textbf{Hyperparameter Sensitivity.}
Finally, we analyzed the sensitivity of the model to two key hyperparameters: the spectral loss weight $\lambda_{\text{freq}}$ and the residual gain factor $\alpha$. Table~\ref{tab:hyperparams} shows that performance is relatively robust around the default values ($\lambda_{\text{freq}}=1.0, \alpha=1.0$). Extreme values (e.g., $\lambda_{\text{freq}}=0.1$ or $\lambda_{\text{freq}}=5.0$) lead to performance degradation, indicating the importance of balancing the generative reconstruction with spectral consistency.

\begin{table}[tbp]
    \scriptsize
    \centering
    \def\arraystretch{1.2}
    \begin{tabular} {c|c|c|c|c|c|c|c}
        \hline
        \multirow{2}{*}{Param} & \multirow{2}{*}{Value} & \multicolumn{3}{c|}{NTU-60\;(Acc, \%)} & \multicolumn{3}{c}{NTU-120 \;(Acc, \%)} \\
        \cmidrule{3-8}
        & & 55/5 & 48/12 & 30/30 & 110/10 & 96/24 & 60/60 \\
        \hline
        \multirow{3}{*}{$\lambda_{\text{freq}}$} 
        & 0.5 & 87.15 & 56.80 & 25.90 & 74.65 & 65.95 & 28.05 \\
        & \textbf{1.0 (Ours)} & \textbf{87.79} & \textbf{57.46} & \textbf{26.55} & \textbf{75.24} & \textbf{66.52} & \textbf{28.67} \\
        & 5.0 & 86.30 & 55.50 & 25.10 & 73.80 & 64.80 & 27.20 \\
        \hline
        \multirow{3}{*}{$\alpha$} 
        & 0.5 & 87.05 & 56.65 & 25.85 & 74.55 & 65.85 & 27.95 \\
        & \textbf{1.0 (Ours)} & \textbf{87.79} & \textbf{57.46} & \textbf{26.55} & \textbf{75.24} & \textbf{66.52} & \textbf{28.67} \\
        & 1.5 & 86.10 & 55.20 & 24.95 & 73.50 & 64.50 & 27.05 \\
        \hline
    \end{tabular}
    \caption{Sensitivity analysis of hyperparameters $\lambda_{\text{freq}}$ and $\alpha$.}
  \label{tab:hyperparams}
\end{table}

\noindent\noindent\textbf{Robustness to Temporal Scale and Sampling Rate.} FDSM is inherently length-agnostic by design. The spectral boundary $M$ is computed as a \textit{relative} ratio $M = \lfloor L/4 \rfloor$, so the DCT is always applied to the available $L$ frames and SG-SRM always supervises the same proportional band (the lowest 25\%) of the spectrum. The key insight is that the primary kinematic features of an action---such as the cadence of a stride or the periodicity of a hand gesture---are consistent in \emph{normalized} temporal coordinates~\cite{cutler2002robust}: they occupy the same relative frequency band regardless of sequence length. For varying temporal scales (e.g., cropping a 300-frame sequence to 150 frames), $L$ decreases and $M$ scales proportionally ($M{=}75 \to 37$), preserving this semantic alignment. For varying sampling rates (e.g., halving the frame rate), the reduced $L$ likewise lowers $M$ proportionally to the \emph{available} bandwidth, preventing the boundary from drifting into the aliased region caused by the reduced Nyquist frequency.

To empirically validate these properties, we evaluate FDSM's robustness via two degradation settings on the NTU-60 test set: (1) \textbf{Cropping} the original 300-frame sequence to shorter contiguous segments (simulating shorter temporal scales), and (2) uniformly \textbf{Downsampling} the sequence (simulating lower sampling rates). As shown in Table~\ref{tab:temporal_robust}, FDSM consistently outperforms TDSM across all conditions, and the results reveal a synergistic dual-level adaptation mechanism.

\textbf{At long temporal scales and high sampling rates} (e.g., Original or $L=150$), the sequences contain rich, fully unfolded high-frequency micro-dynamics. FDSM adapts via its relative spectral boundary $M = \lfloor L/4 \rfloor$: because $M$ scales dynamically with sequence length, SG-SRM can freely isolate and amplify fine-grained details without being constrained by an absolute frame count, yielding consistent improvements over TDSM.

\textbf{At short temporal scales and low sampling rates} (e.g., $L=75$ or 1/4 downsampling), high-frequency visual dynamics are physically compressed or obliterated. Pure-visual alignment methods like TDSM suffer severe degradation (up to $-11.04\%$) as the model has no recourse when the visual signal is impoverished. Under these conditions, FDSM's \textbf{Curriculum Abstraction} module acts as a critical safety net: the kinematically dense semantic priors derived from LLMs compensate for the missing visual dynamics, enabling the model to maintain strong performance even when the motion signal is heavily degraded.

\begin{table}[tbp]
    \scriptsize
    \centering
    \def\arraystretch{1.2}
    
        \begin{tabular}{l|c|cc|cc}
        \hline
        \multirow{2}{*}{Method} & \multirow{2}{*}{Original} & \multicolumn{2}{c|}{Temporal Scales (Cropping)} & \multicolumn{2}{c}{Sampling Rates (Downsampling)} \\
        \cmidrule{3-6}
        & & $L=150$ & $L=75$ & 1/2 & 1/4 \\
        \hline
        TDSM~\cite{do2025bridging} & $86.49$ & $82.70$ & $75.45$ & $83.60$ & $78.15$ \\
        \textbf{FDSM (Ours)} & \textbf{87.79} & \textbf{85.15} & \textbf{79.80} & \textbf{86.10} & \textbf{82.45} \\
        \hline
    \end{tabular}
    \caption{{Robustness to temporal scale and sampling rate on NTU-60 (55/5, Top-1 Acc, \%).}}

    \label{tab:temporal_robust}
\end{table}

\noindent\textbf{Complexity Analysis.} While FDSM introduces three modules, TASL and Curriculum Abstraction are \textit{training-only} strategies that add zero parameters and zero inference latency. The sole structural addition, SG-SRM, is a lightweight component that introduces negligible computational overhead. As shown in Table~\ref{tab:complexity}, FDSM incurs essentially zero deployment penalty ($+0.0001$ GFLOPs, $+0.01$\,s inference time) while achieving substantial performance gains.

\begin{table}[tbp]
    \centering
    \caption{Complexity and performance comparison between TDSM and FDSM. Training time is evaluated on a single NVIDIA L40S GPU for the NTU-60 55/5 split (50{,}000 iterations); inference time is measured per sequence.}
    \label{tab:complexity}
    \scriptsize
    \setlength{\tabcolsep}{3pt}
    \begin{tabular}{lcccccc}
        \toprule
        \multirow{2}{*}{Method} & \multirow{2}{*}{Params\,(M)} & \multirow{2}{*}{GFLOPs} & Training & Inference & \multicolumn{2}{c}{Accuracy\,(\%)} \\
        \cmidrule{6-7}
        & & & Time & Time & 55/5 & 80/40 \\
        \midrule
        TDSM~\cite{do2025bridging} & 261.21 & 6.4545 & 6.7 h & 0.43 s & 86.49 & 36.95 \\
        \textbf{FDSM (Ours)} & \textbf{261.21} & \textbf{6.4546} & \textbf{6.9 h} & \textbf{0.44 s} & \textbf{87.79 (+1.30)} & \textbf{39.16 (+2.21)} \\
        \bottomrule
    \end{tabular}
\end{table}

\begin{figure}[t]
    \centering
    \safeincludegraphics[width=1.0\linewidth]{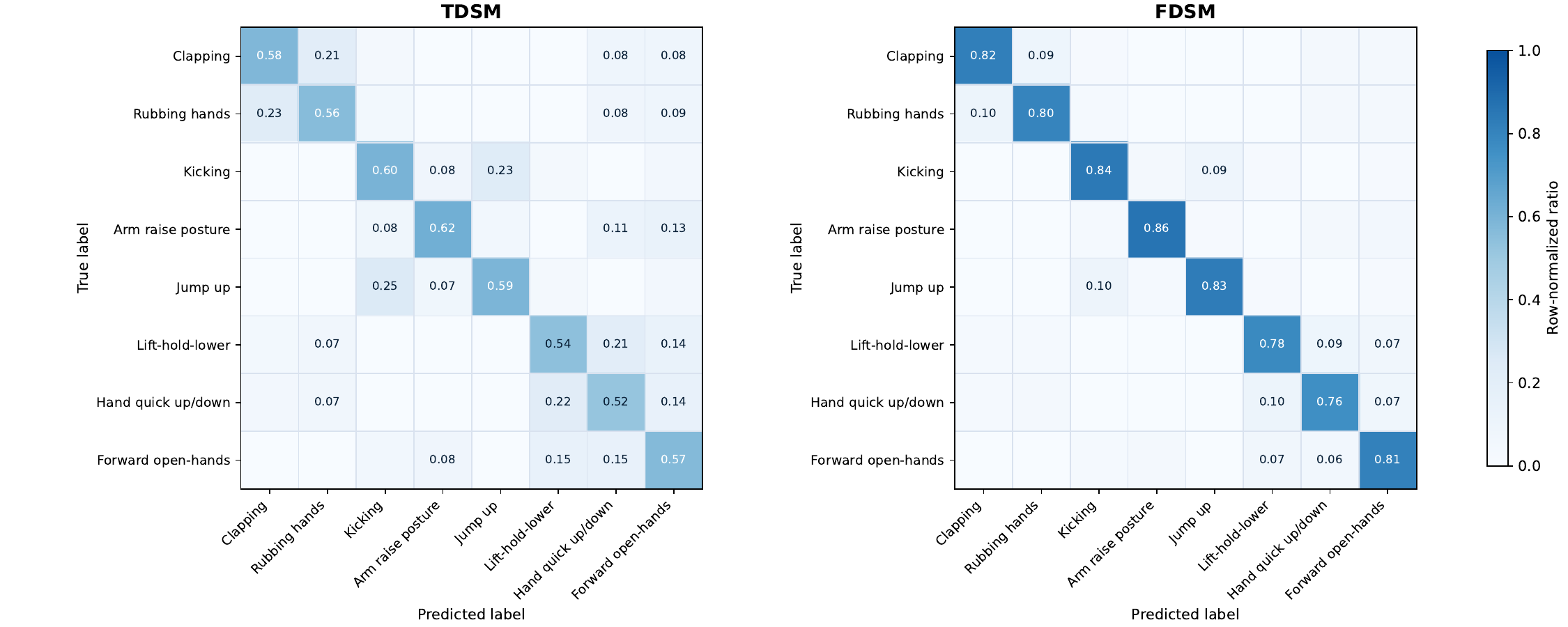}
    \caption{Localized confusion matrix comparison between TDSM (Left) and our FDSM (Right) on selected NTU-120 unseen classes. FDSM significantly reduces the confusion between kinematically similar actions (e.g., ``Clapping'' vs. ``Rubbing hands'') by recovering discriminative high-frequency details.}
    \label{fig:confusion}
\end{figure}

\noindent\noindent\textbf{Qualitative Confusion Analysis.} Fig.~\ref{fig:confusion} presents side-by-side localized confusion matrices for TDSM and FDSM on selected NTU-120 unseen classes. FDSM substantially reduces off-diagonal confusion between kinematically similar pairs (e.g., ``Clapping'' vs.\ ``Rubbing hands''), with corresponding improvements on the diagonal, confirming that recovering high-frequency micro-dynamics is critical for fine-grained action discrimination.

\begin{figure}[t]
    \centering
    \safeincludegraphics[width=\linewidth]{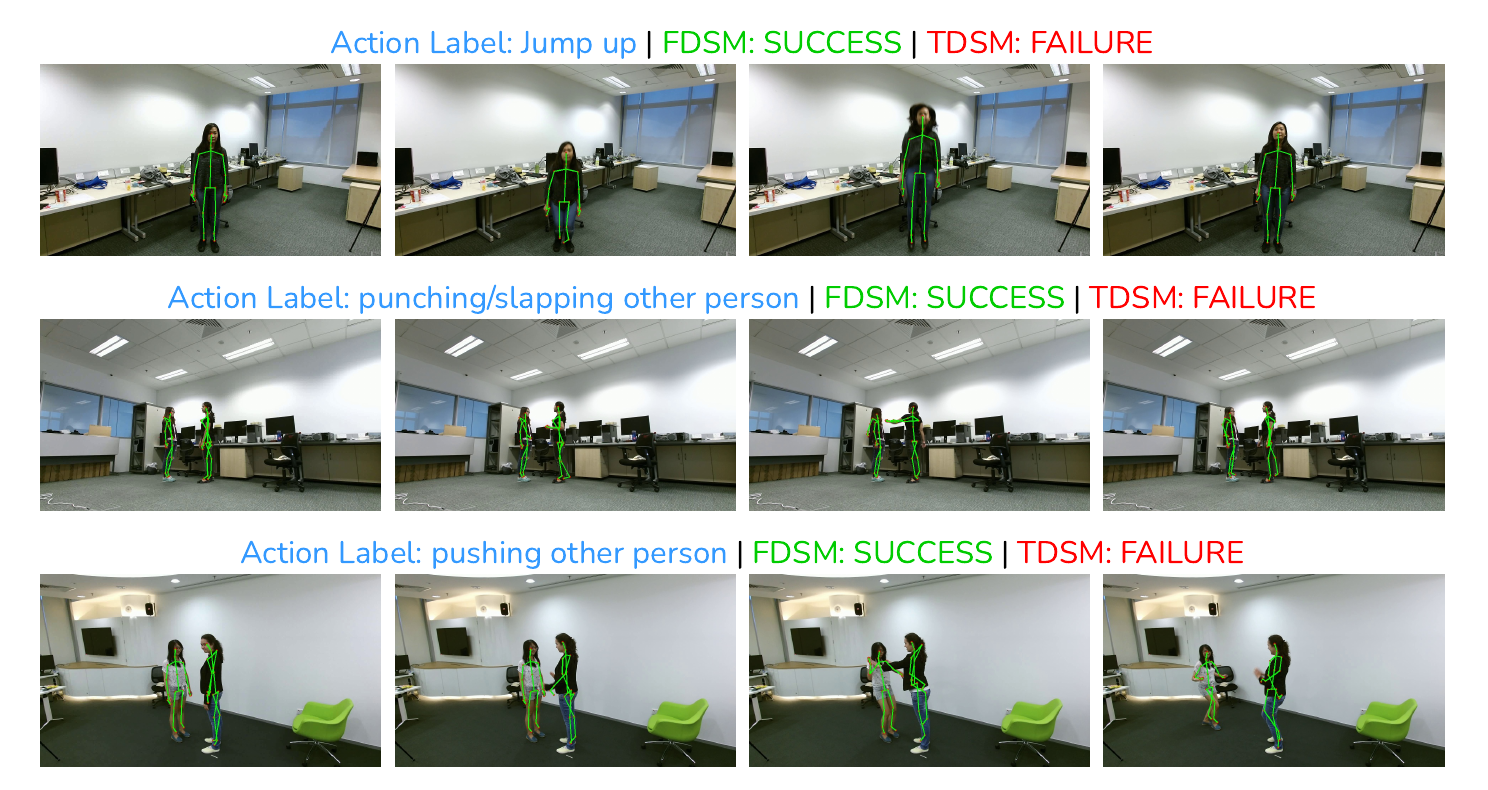}
    \caption{Success cases of FDSM on NTU-120 unseen classes (A027 ``Jump up'', A050 ``Punching/slapping other person'', A052 ``Pushing other person''). Skeletons are overlaid on RGB frames using ground-truth Kinect-captured joint coordinates (colorX/colorY) for visualization; no pose estimator is applied. FDSM correctly classifies all three actions dominated by high-frequency micro-dynamics, while TDSM misclassifies them due to spectral smoothing.}
    \label{fig:success_failure}
\end{figure}

\noindent\noindent\textbf{Success Case Analysis.} Fig.~\ref{fig:success_failure} visualizes representative success cases using skeleton sequences drawn directly from the test set, paired with the classification outcomes of FDSM and TDSM. Three NTU-120 unseen classes dominated by high-frequency micro-dynamics---``Jump up'' (A027), ``Punching/slapping other person'' (A050), and ``Pushing other person'' (A052)---are consistently classified correctly by FDSM but misclassified by TDSM as ``Hopping'', ``Pushing other person'', and ``Punching/slapping other person'', respectively, confirming that SG-SRM's spectral correction recovers the discriminative fine-grained dynamics that the baseline suppresses.

\subsection{LLM Prompt Templates}\label{app:prompts}
This section provides the exact prompt templates used for Large Language Model (LLM) interaction in our framework.
\subsubsection{Rich Kinematic Description Prompt}\label{app:rich_prompt}
The following prompt is used to generate detailed kinematic descriptions for each action class to support the \textit{Curriculum-based Semantic Abstraction} strategy:
\begin{quote}
\textit{``As an expert in human kinesiology and computer vision, please provide [$N_{desc}$] distinct, detailed descriptions for the human action: [Action Name]. Each description should focus on: (1) The specific body parts involved (e.g., wrists, knees, torso); (2) The temporal phases of the movement (e.g., preparation, execution, recovery); and (3) The dynamic characteristics such as speed, rhythm, and intensity. Avoid generic phrases and focus on observable skeletal kinematics.''}
\end{quote}
\subsubsection{Motion Intensity Scoring Prompt}\label{app:intensity_prompt}
The following prompt is used to extract kinematic priors for the \textit{Semantic-Guided Spectral Residual Module}:
\begin{quote}
\textit{``Given the action class [Action Name], output one binary label for motion intensity used in skeleton dynamics: 1 = high-frequency/dynamic (rapid limb transitions, jitter-like fine motion), 0 = low-frequency/static (slow or steady posture-dominant motion). Output only one character: 0 or 1.''}
\end{quote}

\subsection{Action Intensity Statistics}\label{app:intensity_stats}
To provide a comprehensive view of the kinematic distribution across different benchmarks, we analyze the motion intensity labels generated by the LLM for each dataset individually. These labels serve as the supervision for our kinematic projection head. As shown in Table~\ref{tab:intensity_stats}, the distribution remains relatively stable across benchmarks, with a slight shift towards high-intensity actions in the large-scale Kinetics datasets due to their inclusion of diverse sports and complex outdoor activities.

\begin{table}[h]
    \centering
    \label{tab:intensity_stats}
    
        \begin{tabular}{lccc}
            \toprule
            \textbf{Dataset} & \textbf{High Intensity (1)} & \textbf{Low Intensity (0)} & \textbf{Examples (High vs. Low)} \\
            \midrule
            NTU-60~\cite{Ntu60} & 53.3\% (32) & 46.7\% (28) & Punching vs. Reading \\
            NTU-120~\cite{Ntu120} & 55.8\% (67) & 44.2\% (53) & Butt kicks vs. Yawn \\
            PKU-MMD~\cite{PKUMMD} & 54.9\% (28) & 45.1\% (23) & Kicking vs. Bow \\
            Kinetics-200~\cite{yan2018spatial} & 58.0\% (116) & 42.0\% (84) & High jump vs. Drinking \\
            Kinetics-400~\cite{Kinetics} & 58.2\% (233) & 41.8\% (167) & Breakdancing vs. Dining \\
            \bottomrule
        \end{tabular}
        \caption{Detailed distribution of LLM-derived binary motion intensity labels ($s_y^{GT}$) across the five evaluated benchmarks.}
    
\end{table}

\subsection{Selection of Inference Timestep $t_{\text{test}}$}\label{app:ttest_selection}
Unlike iterative generative sampling, our framework performs a one-step noise reconstruction at a fixed timestep $t_{\text{test}}$. To determine the optimal value, we conducted a sensitivity analysis across $t_{\text{test}} \in [0, 50]$ (where $T=50$ is the total diffusion training steps). As shown in Table~\ref{tab:ttest_sensitivity}, $t_{\text{test}}=25$ represents the optimal value for zero-shot inference across multiple benchmarks. At lower $t_{\text{test}}$ (e.g., 10), the skeletal features are insufficiently perturbed by noise, limiting the model's ability to ``re-generate'' discriminative high-frequency details from the semantic prompt. Conversely, at higher $t_{\text{test}}$ (e.g., 40 or 50), the excessive noise level begins to overwrite the fundamental skeletal topology, leading to a loss of global pose coherence. $t_{\text{test}}=25$ provides the ideal balance, allowing the model to correct spectral bias while preserving the fundamental kinematic topology.

\begin{table}[h]
\centering
\label{tab:ttest_sensitivity}
    
        \begin{tabular}{l|cccccc}
        \toprule
        {$t_{\text{test}}$ Value} & {10} & {20} & {\textbf{25 (Ours)}} & {30} & {40} & {50} \\
        \midrule
        {NTU-60 (55/5)} & {84.62} & {86.95} & {\textbf{87.79}} & {86.31} & {82.54} & {72.40} \\
        {NTU-120 (110/10)} & {72.15} & {74.42} & {\textbf{75.24}} & {73.95} & {70.62} & {61.35} \\
        \bottomrule
    \end{tabular}
    \caption{{Zero-shot action recognition accuracy (\%) under varying inference timesteps $t_{\text{test}}$ on NTU-60 and NTU-120 datasets.}}
    
\end{table}

\subsection{Comparison with TDSM}

A central concern in evaluating FDSM is whether it constitutes a genuine architectural advance over TDSM~\cite{do2025bridging} or merely a module-level replacement. We address this by tracing the Spectral Bias to its root cause in TDSM's design and demonstrating how each FDSM component addresses exactly one aspect of it.

TDSM constructs a Zero-Shot Skeleton Action Recognition framework by wrapping a Diffusion Transformer (DiT) around a graph-based latent space, conditioning generation on CLIP-encoded class-name prompts with a uniform $\ell_2$ diffusion loss. This design embeds three coupled structural limitations. First, the DiT's self-attention globally pools spatial and temporal features, making it act as an implicit low-pass filter that systematically suppresses high-frequency temporal dynamics~\cite{si2022inception}. Second, the uniform $\ell_2$ loss concentrates gradient signal on early denoising timesteps that recover only coarse structure, providing no supervision for fine-grained high-frequency detail~\cite{rahaman2019spectral}. Third, bare class-name prompts carry no kinematic specificity---the label ``kicking'' conveys no information about the velocity profile, joint trajectory, or rhythm that define the action's spectral signature. Together, these three choices produce a systematic drop in high-frequency latent energy relative to ground-truth sequences, a deficit directly measurable via DCT-based spectral decomposition and confirmed in Fig.~\ref{fig:psd}.

FDSM targets this single root cause, which is Spectral Bias, by replacing each contributing component with a spectrally aware counterpart. \textbf{SG-SRM} replaces the plain DiT backbone with a DCT-based residual branch gated by a learned kinematic intensity score, directly counteracting the low-pass architectural tendency. \textbf{TASL} replaces the uniform $\ell_2$ loss with timestep-adaptive spectral weighting that concentrates high-frequency supervision at the final denoising steps, where structural fine detail is resolved. \textbf{Curriculum Abstraction} replaces static class-name prompts with a rich-to-sparse LLM-description schedule, endowing the semantic condition with the kinematic specificity the spectral corrections require. The additive ablation in Table~\ref{tab:incremental_ablation} confirms that each component yields an independent, cumulative gain, and that removing any single module reverts performance toward the TDSM baseline---validating that the three components are not interchangeable substitutions but a coherent response to three distinct facets of the same underlying problem.

\begin{table}[h]
    \scriptsize
    \centering
    
    \def\arraystretch{1.2}
    \begin{tabular}{lccc|cc|cc}
        \hline
        \multirow{2}{*}{Methods} & \multirow{2}{*}{SG-SRM} & \multirow{2}{*}{$\mathcal{L}_{\text{Freq}}$} & \multirow{2}{*}{Curr.} & \multicolumn{2}{c|}{NTU-60 (Acc, \%)} & \multicolumn{2}{c}{NTU-120 (Acc, \%)} \\
        \cmidrule{5-8}
        & & & & 55/5 & 30/30 & 110/10 & 60/60 \\
        \hline
        TDSM~\cite{do2025bridging} & & & & 86.49 & 25.88 & 74.15 & 27.21 \\
        + SG-SRM & \checkmark & & & 87.03 & 26.02 & 74.61 & 28.16 \\
        + $\mathcal{L}_{\text{Freq}}$ (TASL) & \checkmark & \checkmark & & 87.31 & 26.11 & 74.89 & 28.32 \\
        \textbf{FDSM (Full)} & \checkmark & \checkmark & \checkmark & \textbf{87.79} & \textbf{26.55} & \textbf{75.24} & \textbf{28.67} \\
        \hline
    \end{tabular}
    \caption{Additive ablation study starting from the TDSM baseline. Each component addresses one distinct facet of the Spectral Bias, yielding cumulative performance gains.}
    \label{tab:incremental_ablation}
    
\end{table}

\subsection{Limitations}

While our framework achieves state-of-the-art performance, a key limitation warrants acknowledgment. The framework inherits the intrinsic stochasticity of diffusion-based inference. Following the protocol established in~\cite{do2025bridging}, our classification relies on sampling random Gaussian noise to probe the generative likelihood of candidate classes. Consequently, the recognition results exhibit minor fluctuations across different noise initializations. Although our extensive experiments confirm that FDSM consistently outperforms baselines regardless of this variability, the non-deterministic nature of the inference process remains a constraint for applications requiring strictly reproducible outputs. Future research could address this by exploring deterministic sampling strategies (e.g., DDIM inversion) or employing ensemble methods to marginalize the noise variance.

\section{Conclusion}
\label{sec:conclusion}

In this work, we identified and addressed the critical limitation of spectral bias in diffusion-based Zero-Shot Skeleton Action Recognition. We argued that standard generative backbones, while effective for global topology, inherently act as low-pass filters that suppress the high-frequency micro-dynamics essential for distinguishing fine-grained actions. 
To overcome this, we proposed a unified {Frequency-Aware Diffusion Framework}. By introducing the {Semantic-Guided Spectral Residual Module}, we endowed the model with the architectural capacity to selectively amplify kinematic details, governed by an internalized frequency prior distilled from LLM knowledge. Complementing this, our {Timestep-Adaptive Spectral Loss} aligned the optimization objective with the intrinsic coarse-to-fine trajectory of the diffusion process, ensuring physically valid reconstruction. Furthermore, our {Curriculum-based Semantic Abstraction} strategy successfully bridged the cognitive gap between sparse labels and complex motion patterns.
Extensive experiments demonstrate that our approach achieves state-of-the-art performance in ZSAR, suggesting that explicit spectral regulation is an important factor in bridging the modality gap. We hope this work inspires further exploration into frequency-aware generative modeling for cross-modal understanding.

\bibliography{sn-bibliography}


\end{document}